\documentclass[twoside]{article}

\usepackage{graphicx}
\usepackage{amsmath}
\usepackage{amssymb}
\usepackage{xspace}
\usepackage{lipsum}  
\usepackage{xcolor}  
\usepackage{acro}  
\usepackage{subcaption}  
\usepackage{booktabs}  
\usepackage{tabularx}  
\usepackage{multirow}  
\usepackage{placeins}  
\usepackage{orcidlink}  

\usepackage{pgfplots}
\pgfplotsset{compat=1.15}
\pgfplotsset{
    discard if not/.style 2 args={
        x filter/.code={
            \edef\tempa{\thisrow{#1}}
            \edef\tempb{#2}
            \ifx\tempa\tempb
            \else
                
            \fi
        }
    }
}
\usepgfplotslibrary{colorbrewer}
\pgfplotsset{
    cycle list/Dark2,
    cycle multiindex* list={
        mark list*\nextlist
        Dark2\nextlist
    },
}
\usetikzlibrary{calc}
\usetikzlibrary{positioning}

\usepackage{hyperref}

\usepackage[accepted]{aistats2022}

\usepackage[round]{natbib}

\newcommand{\bx}{\mathbf{x}}
\newcommand{\by}{\mathbf{y}}
\newcommand{\bz}{\mathbf{z}}

\newcommand{\nE}{\mathbb{E}}

\newcommand{\nR}{\mathbb{R}}

\newcommand{\figref}[1]{Fig.~\ref{#1}}
\newcommand{\secref}[1]{Section~\ref{#1}}

\newcommand{\eqnref}[1]{Eq.~\eqref{#1}}
\newcommand{\tabref}[1]{Table~\ref{#1}}

\makeatletter
\DeclareRobustCommand\onedot{\futurelet\@let@token\@onedot}
\def\@onedot{\ifx\@let@token.\else.\null\fi\xspace}
\def\eg{e.g\onedot} 
\def\ie{i.e\onedot}

\makeatother

\newcommand{\boldparagraph}[1]{\vspace{0.15cm}\noindent{\bf #1:} }

\definecolor{sigma1_l}{HTML}{66c2a5}
\definecolor{sigma2_l}{HTML}{fc8d62}
\definecolor{sigma3_l}{HTML}{8da0cb}

\definecolor{sigma1_d}{HTML}{1b9e77}
\definecolor{sigma2_d}{HTML}{d95f02}
\definecolor{sigma3_d}{HTML}{7570b3}

\DeclareAcronym{acgan}{short=AcGAN, long=Auxiliary Classifier \ac{gan}}
\DeclareAcronym{cd}{short=CD, long=\emph{Chamfer's Distance}}
\DeclareAcronym{cnn}{short=CNN, long=convolutional neural network}
\DeclareAcronym{cov}{short=COV, long=coverage}
\DeclareAcronym{dnn}{short=DNN, long=deep neural network}
\DeclareAcronym{emd}{short=EMD, long=\emph{Earth Mover's Distance}}
\DeclareAcronym{fpd}{short=FPD, long=\emph{Fr\'echet Point Cloud Distance}}
\DeclareAcronym{gan}{short=GAN, long=generative adversarial network}
\DeclareAcronym{jsd}{short=JSD, long=Jensen-Shannon Divergence}
\DeclareAcronym{kde}{short=KDE, long=kernel density estimation}
\DeclareAcronym{knn}{short=$k$-NN, long=$k$-nearest neighbor}
\DeclareAcronym{mmd}{short=MMD, long=minimum matching distance}
\DeclareAcronym{mse}{short=MSE, long=mean squared error}
\DeclareAcronym{vae}{short=VAE, long=variational autoencoder}

\begin{document}

\runningtitle{Point Cloud Generation with Continuous Conditioning}

\runningauthor{Larissa T. Triess, Andre B\"uhler, David Peter, Fabian B. Flohr, J. Marius Z\"ollner}

\twocolumn[

\aistatstitle{Point Cloud Generation with Continuous Conditioning}

\aistatsauthor{
    \textbf{Larissa T. Triess}$^{1,2}$ \orcidlink{0000-0003-0037-8460}
    \quad\quad
    \textbf{Andre B\"uhler}$^{1,3}$
    \quad\quad
    \textbf{David Peter}$^{1}$ \orcidlink{0000-0001-7950-9915}
    \\
    \textbf{Fabian B. Flohr}$^{1,3}$ \orcidlink{0000-0002-1499-3790}
    \quad\quad
    \textbf{J. Marius Z\"ollner}$^{2,4}$ \orcidlink{0000-0001-6190-7202}
    \\
}

\aistatsaddress{
    \\
    $^{1}$Mercedes-Benz AG
    \quad\quad
    $^{2}$Karlsruhe Institute of Technology
    \\
    $^{3}$University of Stuttgart
    \quad\quad
    $^{4}$Research Center for Information Technology
}

]  

\begin{abstract}

Generative models can be used to synthesize 3D objects of high quality and diversity.
However, there is typically no control over the properties of the generated object.
This paper proposes a novel \ac{gan} setup that generates 3D point cloud shapes conditioned on a continuous parameter.
In an exemplary application, we use this to guide the generative process to create a 3D object with a custom-fit shape.
We formulate this generation process in a multi-task setting by using the concept of auxiliary classifier \acp{gan}.
Further, we propose to sample the generator label input for training from a \ac{kde} of the dataset.
Our ablations show that this leads to significant performance increase in regions with few samples.
Extensive quantitative and qualitative experiments show that we gain explicit control over the object dimensions while maintaining good generation quality and diversity.

\end{abstract}

\input{figures/eyecatcher}  

\section{INTRODUCTION}

In recent years many approaches evolved to analyze point clouds, such as classification or segmentation~\citep{Guo2021PCT,Hackel2016CVPR,Li2018NIPS,Qi2017PointNet,Thomas2019ICCV,Wang2019SIGGRAPH,Wu2019CVPR}.
Point clouds are a popular representation for 3D data acquired with depth sensing and laser scanning and are used in many applications, such as automated driving or human-robot interaction.
Here, often complex 3D scenes have to be analyzed by detecting objects, segmenting the scene, or estimating motion~\citep{Zhou2018CVPR,Lang2019CVPR,Liu2019CVPR,Milioto2019IROS,Sirohi2021ArXiv}.
The scenes are composed of a large number of different objects with hierarchical dependencies.
Therefore, most 3D object detection approaches on real-world data use augmentation techniques to improve detection performance.
This is often achieved by placing additional objects into the scene when training the networks~\citep{Yan2018,Lang2019CVPR,Baur2019IV}.
This can greatly improve performance, especially for underrepresented classes, but is limited to the size and diversity of the provided object database.
Therefore, it is favorable to have an unlimited number of objects at ones disposal.

Generative models, such as \acp{gan}~\citep{Goodfellow2014NIPS} or \acp{vae}~\citep{Kingma2014ICLR}, are often used to generate completely new samples with high quality and diversity.
These approaches are initially introduced for image generation, but lately a number of approaches for 3D generation have emerged~\citep{Achlioptas2018ICML,Valsesia2018ICLR,Yang2019PointFlow,Shu2019ICCV,Sun2020WACV,Luo2021CVPR}.
However, none of these methods is capable to actively influence specific properties of the generated object, such as height and width (see~\figref{fig:eyecatcher}).
Therefore, we propose a method to use descriptions in form of continuous conditional parameters within a \ac{gan} to generate objects with desired properties.
These properties can for example describe the aspect-ratios of the object, such that the \ac{gan} generates a custom-fit shape while maintaining generation quality and diversity.
Using continuous conditions introduces new challenges, since it is a mathematically different problem than solving categorical conditioning problems, such as classification~\citep{Ding2021ICLR}.
The challenges arise from an infinitely large set of parameters and conditioning regions where no training data samples exist.

We formulate the continuous conditioning as a multi-task problem within the discriminator.
We add an additional output head, similar to AcGAN~\citep{Odena2017PMLR}, that estimates the continuous parameter.
The overall loss function is then formulated as the weighted sum of the adversarial loss and the regression loss.
Additionally, we propose an alternative label sampling that is used at training time.
In a specifically designed experiment setup, we show that this sampling strategy improves the generation quality of the generator and opens the potential to generate novel out-of-distribution shapes.
To the best of our knowledge, this work is first to use continuous conditional labels for 3D generation.

\section{RELATED WORK}
\label{sec:related}

\subsection{3D Generative Models}

In recent years, a variety of methods emerged from \acp{gan} and \acp{vae} to synthesize realistic data.
Initially proposed to generate realistic 2D images, these concepts have widely been adapted to fulfill a variety of tasks on different modalities~\citep{Song2018ECCV,Chen2019TMM,Karras2019CVPR}.
\cite{Wu2016NIPS}~are first to propose an unsupervised deep generative approach for 3D data generation from probabilistic input.
Their voxel-based \ac{gan} allows to directly adapt concepts from 2D, but is limited in resolution due to computational inefficiency.
Therefore, subsequent works focus on surface~\citep{Mescheder2019CVPR,Chen2019CVPR,Michalkiewicz2019,Park2019CVPR} or point cloud~\citep{Qi2017PointNet,Qi2017PointNet2,Fan2017CVPR} representations instead.
\cite{Achlioptas2018ICML}~propose architectures for both \acp{vae} and \acp{gan} to generate point cloud objects.
The use of relative simple models based on fully connected layers combined with PointNet layers limits the ability to produce more realistic objects.
The method by~\cite{Valsesia2018ICLR} exploits local topology by using a computationally heavy $k$-nearest neighbor technique to produce geometrically accurate point clouds.
\cite{Shu2019ICCV}~aim to improve the expressiveness of the generative models by introducing convolution-like layers and up-sampling techniques to the generator.
\cite{Wang2020}~deal with the choice of suitable discriminator architectures for 3D generation and show that models perform better when focusing on overall object shapes as well as sampling quality.

\subsection{Conditional Generation}

In many cases additional conditioning parameters are desired to generate specific object categories or styles.
The most prominent example, conditional \ac{gan}~\citep{Mirza2014}, uses explicit conditioning and forms the basis of many other approaches~\citep{Zhu2017ICCV,Taigman2017ICLR,Hoffman2019ICML}.
AcGAN~\citep{Odena2017PMLR} refines this concept of class conditioning by using an additional auxiliary classifier in the discriminator to ensure class specific content.
To condition the output on arbitrary combinations of discrete attributes, some works use labeled images as input to the \ac{gan}~\citep{He2019TIP,Perarnau2016NIPSWORK}.
StyleGAN and others investigate how to enhance desirable properties in \ac{gan} latent spaces to influence the characteristics of generated images selectively~\citep{Karras2019CVPR,Karras2020CVPR,Harkonen2020}.

\subsection{Continuous Conditioning}

Many of the aforementioned concepts use discrete label conditioning.
However, attributes like rotation in angles or age in years are by definition continuous.
Using continuous conditions is a mathematically different problem than solving categorical conditioning problems, such as classification~\citep{Ding2021ICLR}.
First, there may be few or no real samples for some regression labels and second, conventional label input methods, \ie one-hot encoding, is not possible for an infinite number of regression labels.
CcGAN~\citep{Ding2021ICLR} is first to introduce a continuous conditional \ac{gan} for image generation.
They solve the aforementioned problems by introducing a new \ac{gan} loss and a novel way to input the labels based on label projection~\citep{Miyato2018ICLR}.
\cite{Shoshan2021}~propose to use attribute specific pre-trained classifiers to enhance desired properties on the generative behavior.
A subsequent training of mapping networks allows to generate noise vectors which produce explicit continuous attributes.
Their method produces convincing results but requires an extensive amount of labeling and well pre-trained classifiers for each attribute.

\subsection{3D Conditional Generation}

The aforementioned conditioning strategies have all been proposed for image synthesis.
There are some works that use text-based conditioning to generate 3D scenes~\citep{Chang2015IJCNLP,Chang2014EMNLP,Chen2019ACCV} with focus on database composition.
Other approaches use symbolic part-tree descriptions to generate 3D objects with predefined compositions~\citep{Mo2020ECCV} or use occupancy networks for image based generation and coloring of 3D objects~\citep{Mescheder2019CVPR}.
Although being related, our method focuses on conditioning point cloud generation using continuous physical parameters.

\subsection{Contributions}

Our main contributions are threefold:
\begin{itemize}
    \item We propose a \ac{gan} setup which formulates the generation process conditioned on continuous parameters in a multi-task setting by adapting the concept of auxiliary classifier \acp{gan}~\citep{Odena2017PMLR}.
    \item We propose to sample the generator conditioning input for training from a \ac{kde} of the parameter distribution. Our ablations show that this leads to a significant performance increase in regions with few samples.
    \item We provide a number of qualitative evaluations that show that we gain explicit control over the object dimensions while maintaining generation quality and diversity.
\end{itemize}

\section{BACKBONE}
\label{sec:treegan}

This section gives a short introduction to our backbone network.
We build upon TreeGAN~\citep{Shu2019ICCV}, a state-of-the-art \ac{gan} architecture that can generate point cloud objects with high quality and diversity.
The generator consists of stacked tree graph convolution layers~(TreeGCN).
It receives a random noise vector~$\bz\in\nR^{96}$ as input and outputs a point cloud~$\bx_\text{gen}=G(\bz)\in\nR^{2048\times3}$.
The loss function of the generator~$G$ is defined as
\begin{equation}
\label{eq:loss_G_adv}
    \mathcal{L}_{G,\text{adv}} =
    - \nE_{\bz \sim \mathcal{Z}} \left[ D\left(\bx_\text{gen}\right) \right] ,
\end{equation}
where $D$ denotes the discriminator.
The latent code distribution~$\mathcal{Z}$ is sampled from a Normal distribution~$\bz\in\mathcal{N}(\mathbf{0}, I)$.

The discriminator follows a PointNet architecture~\citep{Qi2017PointNet}.
It either receives a real~$\bx_\text{real}$ or a generated~$\bx_\text{gen}$ point cloud~$\in\nR^{2048\times3}$ as input and outputs a single scalar~$D(\bx)$.
The output estimates whether the sample originates from the distribution of the real or generated samples.
The loss function of the discriminator~$D$ is defined as
\begin{equation}
\begin{aligned}
\label{eq:loss_D_adv}
    \mathcal{L}_{D,\text{adv}} =
    &\nE_{\bz \sim \mathcal{Z}} \left[ D\left(\bx_\text{gen} \right) \right]
    - \nE_{\bx \sim \mathcal{R}} \left[ D(\bx_\text{real}) \right] \\
    &+ \lambda_\text{gp} \cdot \nE_{\hat{\bx}} \left[ \left( \left\lVert \Delta_{\hat{\bx}} D(\hat{\bx}) \right\rVert_2 - 1 \right)^2 \right] ,
\end{aligned}
\end{equation}
where $\hat{\bx}$ is sampled from an interpolation between real and fake point clouds and $\lambda_\text{gp}$ is the weighting parameter for the gradient penalty term~\citep{Gulrajani2017NIPS}.

\section{METHOD}
\label{sec:method}

\begin{figure}
	\centering

	\resizebox{0.96\linewidth}{!}{\begin{tikzpicture}
\pgfdeclarelayer{bg}    
\pgfsetlayers{bg,main}  
\definecolor{label}{RGB}{123, 185, 199}
\definecolor{network}{RGB}{255, 238, 222}
\definecolor{regression}{RGB}{255, 237, 237}
\tikzset{
    vecinp/.pic = {
        \path[pic actions, draw=black] (-0.1,-0.3) rectangle (0.1,0.3);
        \draw (-0.1, -0.18) -- (0.1, -0.18);
        \draw (-0.1, -0.06) -- (0.1, -0.06);
        \draw (-0.1, 0.06) -- (0.1, 0.06);
        \draw (-0.1, 0.18) -- (0.1, 0.18);

        \node[draw=none, text=black, opacity=1.0, anchor=south] at (0,0.3) {\tikzpictext};
    },
    connsq/.pic = {
        \def\w{0.2};
        \path[draw=black] (-\w/2,-\w/2) rectangle +(\w,\w);
    }
}

\draw[white] (0.8,-1.2) rectangle (10.5,5.0);

\coordinate (preal) at (6.2,2.0);
\coordinate (pfake) at (6.2,1.5);
\coordinate (pdis) at (6.6,1.75);

\coordinate (Generator) at (4.1,0.45);
\coordinate (Discriminator) at ($(pdis)+(1.1,0)$);

\coordinate (xgen) at ($(Generator)+(1.1, 0)$);

\fill (preal) circle (2pt);
\fill (pfake) circle (2pt);
\fill (pdis) circle (2pt);

\draw (pfake) -- (pdis);
\draw[->] ($($(preal)!0.5!(pfake)$)!0.5!(pdis)$) -- +(0,-0.5);
\draw[->] ($($(preal)!0.5!(pfake)$)!0.5!(pdis)$) -- +(0,0.5);
\draw[->] (pdis) -- ($(Discriminator)-(0.5,0)$) node[anchor=south, pos=0.5] {$\bx$};

\begin{scope}[shift={(1.4,0.15)}]
    \begin{scope}[shift={(0,0.8)}]
        \begin{scope}[shift={(0.2,0)}]
            \draw[fill=label, fill opacity=0.2] (-0.2, 0.0, 0.0) -- +(0.4, 0, 0) -- +(0.4, 0.4, 0) -- +(0, 0.4, 0) -- cycle;
            \draw[fill=label, fill opacity=0.2] (0.2, 0.0, 0.0) -- +(0, 0, -0.4) -- +(0, 0.4, -0.4) -- +(0, 0.4, 0) -- cycle;
            \draw[fill=label, fill opacity=0.2] (0.2, 0.4, 0.0) -- +(0, 0, -0.4) -- +(-0.4, 0, -0.4) -- +(-0.4, 0, 0) -- cycle;

            \draw[stealth-stealth] (-0.2, -0.1, 0.0) -- +(0.4, 0, 0);
            \draw[stealth-stealth] (0.3, 0.0, 0.0) -- +(0, 0, -0.4);
            \draw[stealth-stealth] (0.3, 0.0, -0.4) -- +(0, 0.4, 0);

            \node[draw=none, anchor=north] at (0,-0.1) {condition};
        \end{scope}

        \coordinate (yhat) at (1.6, 0.3);
        \draw[->] (0.8,0.3) -- ($(yhat)-(0.1,0)$);
        \pic[fill = label, pic text = {$\by_\text{cond}$}] at (yhat) {vecinp};
    \end{scope}

    \begin{scope}[shift={(0,-0.8)}]
        \begin{scope}[scale=0.5,shift={(0.4,0)}]
            \draw[domain=-1:1, smooth, variable=\x, black]  plot ({\x}, {1/(0.4*sqrt(2*pi))*exp(-((\x-0.0)^2)/(2*0.4^2))});
            \draw[->] (-1.4,0) -- (1.4,0);
            \draw[->] (0,-0.2) -- (0,1.4);
            \node[draw=none, anchor=north] at (0,-0.1) {noise};
        \end{scope}

        \coordinate (z) at (1.6,0.3);
        \draw[->] (0.8,0.3) -- ($(z)-(0.1,0)$);
        \pic[fill = white!80!black, pic text = {$\bz$}] at (z) {vecinp};
    \end{scope}

\end{scope}

\draw[->] ($(yhat)+(0.1, 0)$) -| +(0.2, 0) |- ($(Generator)-(0.5,-0.14)$);
\draw[->] ($(z)+(0.1, 0)$) -| +(0.2, 0) |- ($(Generator)-(0.5,0.14)$);

\begin{scope}[shift={(Generator)}]
    \fill[network, draw=black] (-0.5,-0.35) -- (0.5,-0.6) -- (0.5,0.6) -- (-0.5,0.35) -- cycle;
    \node[draw=none, anchor=center] at (0,0) {G};
\end{scope}

\pic [fill = white!80!black, pic text = {$\bx_\text{gen}$}] at (xgen) {vecinp};

\draw[->] ($(Generator)+(0.5,0)$) -- ($(xgen)-(0.1,0)$);  
\draw ($(xgen)+(0.1,0)$) -| ($(pfake)-(0.55,0)$) |- (pfake);  

\coordinate (x) at ($(yhat)+(0,1.6)$);
\coordinate (y) at ($(x)+(0,1.2)$);

\pic [fill = white!80!black, pic text = {$\bx_\text{real}$}] at (x) {vecinp};
\pic [fill = label, pic text = {$\by_\text{real}$}] at (y) {vecinp};
\draw[dashed] ($(x)-(0.36,0.5)$) rectangle ($(y)+(0.36,0.8)$);
\node[rotate=90] at ($(x)!0.5!(y)-(0.6,0)$) {dataset};

\draw ($(x)+(0.1,0)$) -| ($(preal)-(0.55,0)$) |- (preal);

\begin{scope}[shift={(Discriminator)}]
    \fill[network, draw=black] (-0.5,-0.6) -- (0.5,-0.35) -- (0.5,0.35) -- (-0.5,0.6) -- cycle;
    \node[draw=none, anchor=center] at (0,0) {D};
\end{scope}

\draw[->] ($(Discriminator)+(0.5,-0.14)$) -- +(1, 0) node[anchor=north, pos=0.5] {$D(\bx)$} |- +(1.2, -0.4) node[anchor=west, draw=black, minimum width=1cm] {$\mathcal{L}_\text{adv}$};

\draw[->] ($(Discriminator)+(0.5,0.14)$) -- +(1, 0) node[anchor=south, pos=0.5] {$\hat{\by}(\bx)$} |- +(1.2, 0.4) node[anchor=west, draw=black, minimum width=1cm] (regloss) {$\mathcal{L}_\text{reg}$};
\draw[<-] (regloss) -- ($(regloss)+(0,0.6)$) node[anchor=south] {$\by$};

\end{tikzpicture}}

	\caption{
		\textbf{Architecture}:
		The generator~$G$ generates a point cloud~$\bx_\text{gen}$ from a random vector~$\bz$ and a continuous parameter~$\by_\text{cond}$.
		It is then alternately -- with a real point cloud~$\bx_\text{real}$ -- fed to the discriminator, which predicts the probability of the sample stemming from the real distribution~$D(\bx)$ and an estimate of the parameter~$\hat{\by}(\bx)$.
		With these two outputs, the adversarial~$\mathcal{L}_\text{adv}$ and regression~$\mathcal{L}_\text{reg}$ losses are computed.
	}

	\label{fig:acgan}
\end{figure}
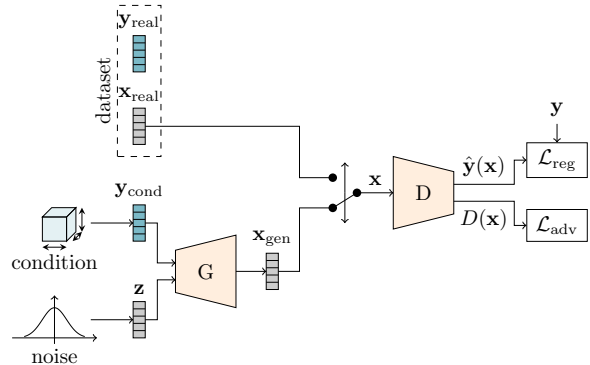

\figref{fig:acgan}~presents the \ac{gan} setup of our proposed continuous conditioning architecture.
The concept for the continuous conditioning is introduced in~\secref{sec:method_labels} while \secref{sec:method_sampling} explains the required parameter sampling
\secref{sec:method_model}~describes the model architecture and the training losses are introduced in~\secref{sec:method_losses}.

\subsection{Continuous Parameters}
\label{sec:method_labels}

We aim to control the outer dimensions of the generated object with the continuous parameters while maintaining diversity regarding the type and shape of the object.
While the object size and shape shall be disentangled, our method is still allowed to modify the shape slightly to better match the defined bounding box in terms of creating realistic and valid shapes, \ie a small table can have a single leg, while a long table typically does not.
We later refer to this property as \emph{semantically meaningful}.
The object is defined as a vector of points~$\bx \in \nR^{N \times 3}$ with the number of points~$N=2048$ and the dimensions~$[x,y,z]$.
The parameter~$\by \in \nR^3$ defines the extent of the object in each dimension $\by\!~\!=\!~\!(\Delta x, \Delta y, \Delta z) \in [0, 1]^3$.
Therefore, the parameters~$\by_\text{real}$ for the training data~$\bx_\text{real}$ can easily be computed from the data itself with~$\by_\text{real}\!~\!=\!~\!\lVert \max\left(\bx_\text{real}\right) - \min\left(\bx_\text{real}\right) \rVert$.

\subsection{Label Sampling for Training}
\label{sec:method_sampling}

At training time, we have to sample parameters~$\by_\text{cond}$ as a second input to the generator~$G$.
The easiest method is to sample randomly in~$[0, 1]$.
However, this can lead to a description of unsuitable object dimensions.
To circumvent this issue, we can sample the dimensions from the training dataset.
This ensures that only actually possible conditioning parameters are sampled.
However, this limits the generator training to a fixed number of input conditioning.
Therefore, we propose to sample the dimensions from the distribution of the dimensions within the training dataset.
We compute a \ac{kde} over the whole dataset prior to training.
This again assures that only suitable dimensions are drawn, but it does not limit them to be present in the dataset.
In all cases, the aim of the generator is to generate point clouds that have dimensions which are close to the conditioned dimensions~$\by_\text{cond}$.

\subsection{Model}
\label{sec:method_model}

\begin{figure}
	\centering

	\begin{subfigure}[t]{0.48\linewidth}
		\centering
		\resizebox{\linewidth}{!}{\begin{tikzpicture}[
    plus/.style={path picture={
        \draw[black] (path picture bounding box.south) -- (path picture bounding box.north) (path picture bounding box.west) -- (path picture bounding box.east);
    }}
]
\definecolor{label}{RGB}{123, 185, 199}
\definecolor{network}{RGB}{255, 238, 222}
\definecolor{tree}{RGB}{125, 173, 250}
\definecolor{fc}{RGB}{116, 219, 173}
\tikzset{
    vecinp/.pic = {
        \path[pic actions, draw=black] (-0.1,-0.3) rectangle (0.1,0.3);
        \draw (-0.1, -0.18) -- (0.1, -0.18);
        \draw (-0.1, -0.06) -- (0.1, -0.06);
        \draw (-0.1, 0.06) -- (0.1, 0.06);
        \draw (-0.1, 0.18) -- (0.1, 0.18);

        \node[draw=none, text=black, opacity=1.0, anchor=south] at (0,0.3) {\tikzpictext};
    },
}

\draw[white] (-1.2, -2) rectangle (4.6, 2);

\fill[network, draw=black] (0,-1.2) -- (3.2,-1.8) -- (3.2,1.8) -- (0,1.2) -- cycle;

\coordinate (y) at (-0.5, 0.55);
\coordinate (z) at (-0.5, -0.55);
\coordinate (fcy) at (0.5, 0.55);
\coordinate (fcz) at (0.5, -0.55);
\coordinate (vin) at (1.7, 0);

\pic[fill = label, pic text = {$\by_\text{cond}$}] at (y) {vecinp};
\pic[fill = white!80!black, pic text = {$\bz$}] at (z) {vecinp};

\fill[fc, draw=black] ($(fcy)-(0.25,0.5)$) rectangle +(0.5, 1) node[text=black, pos=.5, rotate=90] (lineary) {FC};
\fill[fc, draw=black] ($(fcz)-(0.25,0.5)$) rectangle +(0.5, 1) node[text=black, pos=.5, rotate=90] (linearz) {FC};

\node[draw=black, fill=white, rotate=90] (concat) at (1.15, 0) {concat};

\pic[fill = label!70!black] at ($(vin)+(0, 0.3)$) {vecinp};
\pic[fill = white!50!black] at ($(vin)-(0, 0.3)$) {vecinp};

\fill[tree, draw=black] (2, -1.2) rectangle +(1, 2.4) node[text=black, pos=.5, text width=0.8cm, align=center] (tree) {Tree GCN};

\pic[fill = white!80!black, pic text = {$\bx_\text{gen}$}] at (3.7, 0) {vecinp};

\draw[->] ($(y)+(0.1,0)$) -- (lineary.north);
\draw[->] (lineary.south) -| ($(concat.north)-(0.1, -0.2)$) |- ($(concat.north)-(0, -0.2)$);

\draw[->] ($(z)+(0.1,0)$) -- (linearz.north);
\draw[->] (linearz.south) -| ($(concat.north)-(0.1, 0.2)$) |- ($(concat.north)-(0, 0.2)$);

\draw[->] (concat) -- ($(vin)-(0.1,0)$);
\draw[->] ($(vin)+(0.1,0)$) -- (tree.west);
\draw[->] (tree.east) -- (3.6, 0);

\end{tikzpicture}}
		\caption{\label{fig:acgan_generator}Generator}
	\end{subfigure}%
	\hspace{0.5em}%
	\begin{subfigure}[t]{0.48\linewidth}
		\centering
		\resizebox{\linewidth}{!}{\begin{tikzpicture}
\definecolor{label}{RGB}{123, 185, 199}
\definecolor{network}{RGB}{255, 238, 222}
\definecolor{pointnet}{RGB}{125, 173, 250}
\definecolor{fc}{RGB}{116, 219, 173}
\tikzset{
    vecinp/.pic = {
        \path[pic actions, draw=black] (-0.1,-0.3) rectangle (0.1,0.3);
        \draw (-0.1, -0.18) -- (0.1, -0.18);
        \draw (-0.1, -0.06) -- (0.1, -0.06);
        \draw (-0.1, 0.06) -- (0.1, 0.06);
        \draw (-0.1, 0.18) -- (0.1, 0.18);

        \node[draw=none, text=black, opacity=1.0, anchor=south] at (0,0.3) {\tikzpictext};
    },
}

\draw[white] (-1.2, -2) rectangle (4.6, 2);

\fill[network, draw=black] (0,-1.8) -- (3.2,-1.2) -- (3.2,1.2) -- (0,1.8) -- cycle;

\pic[fill = white!80!black, pic text = {$\bx$}] at (-0.5, 0) {vecinp};

\fill[pointnet, draw=black] (0.4, -1.2) rectangle +(1, 2.4) node[text=black, pos=.5, text width=0.8cm, align=center] (pointnet) {Point Net};

\pic[fill = white!60!black] at (1.9, 0) {vecinp};

\fill[fc, draw=black] (2.4, 0.1) rectangle +(0.5, 1) node[text=black, pos=.5, rotate=90] (FC_top) {FC};
\fill[fc, draw=black] (2.4, -1.1) rectangle +(0.5, 1) node[text=black, pos=.5, rotate=90] (FC_bottom) {FC};

\draw[->] (-0.4, 0) -- (pointnet.west);
\draw[->] (pointnet.east) -- (1.8, 0);

\draw[->] (2.0, 0) -| +(0.1, 0) |- (FC_top.north);
\draw[->] (2.0, 0) -| +(0.1, 0) |- (FC_bottom.north);

\draw[->] (FC_top.south) -- +(0.6, 0) node[anchor=west] {$\hat{\by}(\bx)$};
\draw[->] (FC_bottom.south) -- +(0.6, 0) node[anchor=west] {$D(\bx)$};

\end{tikzpicture}}
		\caption{\label{fig:acgan_discriminator}Discriminator}
	\end{subfigure}

	\caption{
		\textbf{Model details}:
		(\subref{fig:acgan_generator})~shows the label input of the generator.
		Both inputs are fed through a fully-connected~(FC) layer, concatenated and then fed to the tree graph convolution network~(TreeGCN).
		(\subref{fig:acgan_discriminator})~shows the discriminator.
		After a common feature extractor~(PointNet), the model splits in two identical parts, where $D(\bx)$ is the adversarial feedback and $\hat{\by}(\bx)$ is the prediction of the continuous object description.
	}

	\label{fig:acgan_details}
\end{figure}
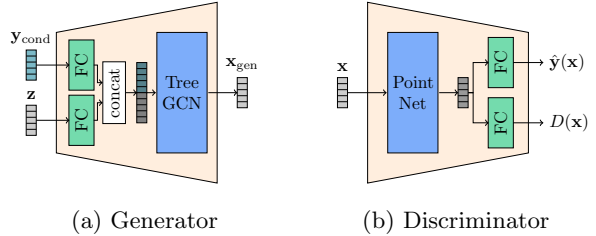

This section only describes the changes made to our backbone architecture for which we use TreeGAN, as described in~\secref{sec:treegan}.
\figref{fig:acgan_details}~shows how the label conditioning is incorporated into the generator and discriminator models.
For the generator (\figref{fig:acgan_generator}), we extend the architecture to receive an additional input, the conditioning vector~$\by_\text{cond}$.
This vector and the noise vector~$\bz \in \nR^{96}$ are both passed through a linear layer.
The concatenated results are then fed to the otherwise unmodified TreeGCN~\citep{Shu2019ICCV} which outputs the point cloud~$\bx_\text{gen}=G(\by_\text{cond},\bz)$.
The discriminator input~$\bx$ is either a point cloud from the dataset~$\bx_\text{real}$ or a generated point cloud from the generator~$\bx_\text{gen}$~(\figref{fig:acgan_discriminator}).
The network has two outputs, one for the standard adversarial feedback~$D(\bx)$ and one to estimate the continuous parameter conditioning of the presented data~$\hat{\by}=\hat{\by}(\bx)$.
This concept is adapted from \ac{acgan}~\citep{Odena2017PMLR,Atienza2019CVPRWORK}.
The idea is to leverage potential synergies between the two tasks within the shared discriminator layers.

\subsection{Losses}
\label{sec:method_losses}

The training objective is composed of two parts, the adversarial loss~$\mathcal{L}_\text{adv}$ and the regression loss~$\mathcal{L}_\text{reg}$ for the continuous parameter.
Just like TreeGAN, we utilize the Wasserstein objective function with gradient penalty for the adversarial loss~\citep{Arjovsky2017,Gulrajani2017NIPS}.
For the parameter regression, we use L2-norm
\begin{equation}
    \mathcal{L}_\text{reg}\left(\by,\hat{\by}\right) = \left\lVert \by - \hat{\by} \right\rVert_2
\end{equation}
with the predicted label~$\hat{\by}$ and its target label~$\by$.

The adversarial part of the generator loss~$\mathcal{L}_{G,\text{adv}}$ is defined in \eqnref{eq:loss_G_adv} with a slight change to the definition of the generated point cloud~$\bx_\text{gen}$.
It is now also dependent on the input conditioning and is therefore defined as $G(\by_\text{cond},\bz)$ (instead of the unconditioned version~$G(\bz)$).
For the generator regression, we compute the $\mathcal{L}_\text{reg}$ between the discriminator prediction for the condition parameter of the generated sample~$\hat{\by}_\text{gen}$ and the actually requested parameter~$\by_\text{cond}$.
This results in the overall generator loss
\begin{equation}
\label{eq:loss_G}
    \mathcal{L}_G = \lambda_\text{adv} \cdot \mathcal{L}_{G,\text{adv}} + \lambda_\text{reg} \cdot \mathcal{L}_\text{reg}\left(\by_\text{cond},\hat{\by}_\text{gen}\right)
\end{equation}
with the loss weighting factors $\lambda_\text{adv}$ and $\lambda_\text{reg}$.

\eqnref{eq:loss_D_adv} defines the adversarial loss of the discriminator.
The regression loss is defined as the error between the parameter of the real data~$\by_\text{real}$ and its corresponding prediction~$\hat{\by}_\text{real}$.
Analogously to the generator, this leads to the discriminator loss
\begin{equation}
\label{eq:loss_D}
    \mathcal{L}_D = \lambda_\text{adv} \cdot \mathcal{L}_{D,\text{adv}} + \lambda_\text{reg} \cdot \mathcal{L}_\text{reg}\left(\by_\text{real},\hat{\by}_\text{real}\right)
\end{equation}
with the same loss weights $\lambda_\text{adv}$ and $\lambda_\text{reg}$ as in~\eqnref{eq:loss_G}.

\section{EXPERIMENTS}
\label{sec:experiments}

This section gives an overview on the experiments with the results reported in~\secref{sec:results}.
Further details are provided in the supplementary material.

\subsection{Dataset and Metrics}
\label{sec:experiments_dataset_metric}

For our experiments, we use ShapeNetPart~\citep{ShapeNetPart2016}, a dataset with part annotations of more than 30,000 3D shapes in 16 object categories from ShapeNetCore~\citep{ShapeNet2015}.
To compare our results in terms of generation quality, we use a pre-trained version of the original \ac{fpd}~\citep{Shu2019ICCV}.
The adherence of the conditioning properties are evaluated by calculating the \ac{mse} for each dimension extent ($\Delta x$, $\Delta y$, $\Delta z$) of the generated object~$\bx_\text{gen}$ versus the desired input parametrization~$\by_\text{cond}$.
These are the two most important metrics, but we also report \ac{cov}, \ac{mmd}, and \ac{jsd} to make our work comparable to existing methods~\citep{Achlioptas2018ICML,Shu2019ICCV}.
The evaluation also focuses on qualitative results to show the advantages of our method.

\subsection{Implementation Details}
\label{sec:experiments_implementation}

We build our model upon the existing implementation of TreeGAN~\citep{Shu2019ICCV} that we further refer to as our backbone model.
The training parameters are also identical.
We only apply the changes introduced in~\secref{sec:method} to keep our method directly comparable to the backbone and do not incorporate any further mechanisms to enhance generation quality.

The loss weights $\lambda_\text{adv}$ and $\lambda_\text{reg}$ from \eqnref{eq:loss_G} and \eqnref{eq:loss_D} are variable and learned together with the rest of the model parameters, as proposed by~\cite{KendallCVPR2018}.

We train all networks for 3000 epochs and select the checkpoint with the lowest combined metric score.
The combined metric is defined as the product of FPD and MSE.
This ensures fidelity as well as correctness and considers differing value ranges.

\begin{figure}
	\centering
	\resizebox{0.9\linewidth}{!}{\input{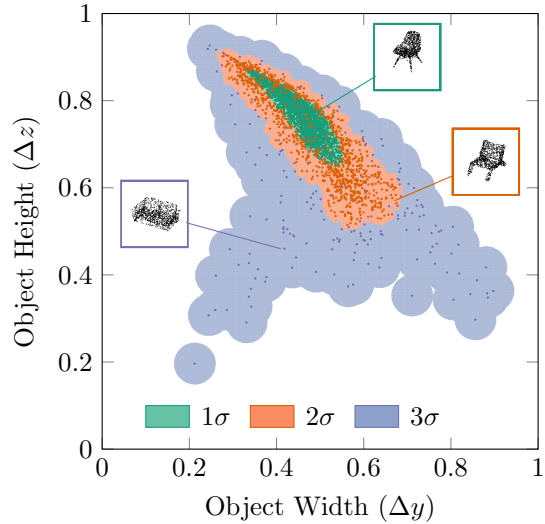}}
	\caption{
		\textbf{Region-classified dimension distribution}:
		The plot shows all samples (marks) of the chair class for the entire training dataset in terms of their dimension extend in height~$z$ and width~$y$.
		The three colors represent the resulting regions from \ac{knn} classifier with $k$=20 based on a \ac{kde}.
		The regions correspond to $1\sigma\approx68\%$ (green), $2\sigma\approx27\%$ (orange), and $3\sigma=5\%$ (blue) of the entire data distribution.
	}
	\label{fig:dimension_distribution}
\end{figure}

\subsection{Baselines}
\label{sec:experiments_baselines}

We compare our method to two baselines.
These baselines represent possible design choices to achieve the desired aim of generating point clouds of specific dimensions when no explicit method already exists.

\boldparagraph{B1 -- Backbone with re-sampling}
A simple method to generate shapes of desired dimensions with an existing model, such as the backbone, is to generate point clouds from multiple sampled $z$~vectors and then choose the one that has the dimensions that are closest to the ones requested.
Here it is expected that the fidelity of the generated samples is similar to the ones from the backbone network itself.
However, the sampling process is time consuming and the dimensions might not exactly fit, especially if these dimensions are not well reflected in the training dataset.
For our experiments, we sample 10 versions per object and select the one with the smallest \ac{mse}.

\boldparagraph{B2 -- Backbone with scaling}
Another variant is to add a subsequent scaling block that squashes the generated point cloud into the desired dimensions.
In contrast to B1, it is expected that the regression error on the dimensions is zero.
However, the generation fidelity can be severely impacted.

\subsection{Distribution Sampling}
\label{sec:experiments_distribution}

A key aspect of our method is that it is possible to actively sample from different regions of the conditioning vector distribution.
Therefore, we specifically investigate the generation capabilities within different regions of the data distribution.
\figref{fig:dimension_distribution}~shows the distribution of the object dimensions of the training dataset for the chair class.
The distribution is divided into three sections, indicated by the coloring.
The sections are classified into regions where $1\sigma\approx68\%$, $2\sigma\approx27\%$, and $3\sigma\approx5\%$, of the data lies, computed with a \ac{kde}.
In additional experiments, we generate 1000 samples for each region and then compare \ac{fpd} and \ac{mse} to show the effectiveness of our method.

\section{RESULTS}
\label{sec:results}

\subsection{Quantitative Results}
\label{sec:results_quantitative}

\tabref{tab:quantitative_results}~contains quantitative comparisons in terms of typical metrics used to evaluate the quality of generated point cloud objects.
The rows for the backbone are included for reference but cannot be used for comparison, as it does not have the conditioning ability.
B1 and B2 each symbolize the two corner cases.
On the one hand, B1 obtains the lowest \ac{fpd} but the highest \ac{mse}.
This is the result of exploiting the good generation quality of the backbone and combine it with a sampling mechanism to obtain object of desired dimensions.
However, it becomes clear that it is hard to obtain the desired shape configuration in an acceptable inference time, while the inference time depends on how many samples have to be drawn for one inference step.
On the other hand, there is B2 which obtains an \ac{mse} of zero by construction.
It simply scales the object to the desired size.
However, this comes at the cost of realism, as evident from the increase in \ac{fpd}.
Our proposed approach lies in between B1 and B2 regarding \ac{fpd} and obtains a very low \ac{mse} of only $0.28\%$.
The other metrics show relatively equal performance for all methods.

The numbers indicate that our method is capable to ensure desired dimensions while maintaining high quality.
However, we want to stress that the full potential of our method is better visible in the following qualitative results.

It is important to note that these baselines are only applicable since it is possible to easily compute the dimension parameter from the data itself.
For many other applications this is not the case and our method offers the suitable solution (see also~\secref{sec:discussion}).

Additionally to the presented results, we also evaluated the label incorporation of traditional cGAN~\citep{Mirza2014} and CcGAN~\citep{Ding2021ICLR} combined with the backbone network.
However, these methods did not achieve satisfying results in our setup.
In particular, cGAN completely ignored the conditioning and therefore received \ac{mse} values over 50\% (at \ac{fpd}$\approx1.9$ for ``chair'').
While for CcGAN, training was unstable and therefore resulted in metric scores that are several magnitudes higher than usual.
For more information and additional results, we refer to the supplemental material.

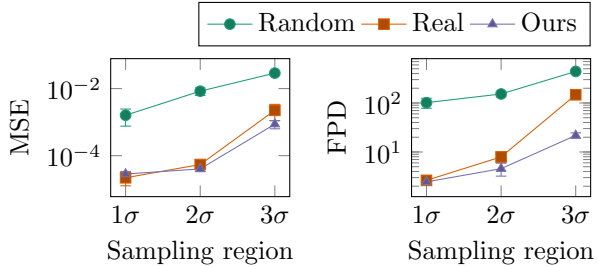
\begin{figure}
	\centering

    \begin{tikzpicture}
        \begin{semilogyaxis}[
            width=0.48\linewidth,
			ymin=6e-6,
            ylabel={MSE},
            xlabel={Sampling region},
            symbolic x coords={1,2,3,4},
            xtick=data,
            xticklabels={$1\sigma$,$2\sigma$,$3\sigma$},
        ]
            \foreach \i in {2,3,4}{
                \addplot+[discard if not={x}{\i},error bars/.cd,y dir=both,y explicit]
                table [x=sigma, y expr=\thisrow{mse_mean}, y error=mse_std, col sep=comma]
                {data/distribution_sampling.csv};
            }
        \end{semilogyaxis}

        \begin{semilogyaxis}[
			xshift=4cm,
            width=0.48\linewidth,
            ylabel={FPD},
            xlabel={Sampling region},
            symbolic x coords={1,2,3,4},
            xtick=data,
            legend style={at={(1,1.25)}, anchor=east},
			legend columns=-1,
            xticklabels={$1\sigma$,$2\sigma$,$3\sigma$},
        ]
            \foreach \i in {2,3,4}{
                \addplot+[discard if not={x}{\i},error bars/.cd,y dir=both,y explicit]
                table [x=sigma, y=fpd_mean, y error=fpd_std, col sep=comma]
                {data/distribution_sampling.csv};
            }
            
            \legend{Random,Real,Ours}
        \end{semilogyaxis}
    \end{tikzpicture}

	\caption{
		\textbf{Distribution sampling performance}:
		The figure shows performances for three data sampling regions for our model trained with three different label sampling strategies.
		The regions correspond to the labels on the x-axis, as defined in~\figref{fig:dimension_distribution}.
		For each region, 1000 samples are generated.
		The left plot shows the \ac{mse} of the dimension regression.
		On the right, the \ac{fpd} of the generated samples is presented.
	}
	\label{fig:distribution_sampling}
\end{figure}  
\begin{table*}
    \centering
    \caption{
        \textbf{Quantitative Comparison}:
        We report results for the classes ``Chair'' and ``Airplane'' in terms of the metrics used by~\cite{Shu2019ICCV}.
        Additionally, we report the regression error (MSE) for our introduced task.
        We report the results for the backbone network (TreeGAN) for reference.
        All evaluations are conducted on a hold-out validation split.
        \cite{Shu2019ICCV}~use the entire dataset for training, therefore values vary slightly.
        Best results are marked in \textit{bold}, second best in \textit{italic}.
    }
    \label{tab:quantitative_results}
    \begin{tabularx}{\linewidth}{l X c c cc c cc c}
        \toprule

        \multirow{2}{*}{Shape} &
        \multirow{2}{*}{Model} &
        \multirow{2}{*}{FPD ($\downarrow$)} &
        \multirow{2}{*}{MSE [\%] ($\downarrow$)} &
        \multicolumn{2}{c}{MMD ($\downarrow$)} &&
        \multicolumn{2}{c}{COV ($\uparrow$)} &
        \multirow{2}{*}{JSD ($\downarrow$)} \\
        \cline{5-6}\cline{8-9}
        & & & & CD & EMD && CD & EMD & \\

        \midrule

        \multirow{5}{*}{Chair}
        & Backbone & 0.9525 & -- & 0.0020 & 0.1027 && 0.4875 & 0.2500 & 0.1082 \\
        \cline{2-10}
        & Baseline~1 (B1) & \textbf{1.3674} & 26.07 & 0.0023 & 0.1013 && 0.4750 & 0.2625 & 0.1123  \\
        & Baseline~2 (B2) & 1.9259 & \textbf{0.00} & 0.0021 & 0.1003 && 0.4875 & 0.2625 & 0.1068 \\
        & Ours & \textit{1.5290} & \textit{0.28} & 0.0022 & 0.1059 && 0.4625 & 0.3125 & 0.1434 \\

        \midrule

        \multirow{5}{*}{Airplane}
        & Backbone & 1.2947 & -- & 0.0002 & 0.0805 && 0.4375 & 0.1375 & 0.1887 \\
        \cline{2-10}
        & Baseline~1 (B1) & \textit{1.0209} & 15.08 & 0.0003 & 0.0812 && 0.4500 & 0.1125 & 0.1819 \\
        & Baseline~2 (B2) & 1.6613 & \textbf{0.00} & 0.0003 & 0.0783 && 0.5250 & 0.1375 & 0.1834 \\
        & Ours & \textbf{0.8691} & \textit{0.30} & 0.0003 & 0.0724 && 0.5000 & 0.1250 & 0.1291 \\

        \bottomrule
    \end{tabularx}
\end{table*}

\subsection{Label and Region Sampling Ablations}
\label{sec:results_sampling}

\figref{fig:distribution_sampling}~shows the results of our distribution sampling experiments on the chair class (explained in~\secref{sec:experiments_distribution}).
We report results for three differently trained versions of our model.
They differ in the way the conditioning parameter~$\by_\text{cond}$ is sampled at training time (see~\secref{sec:method_sampling}).
The green dots correspond to random sampling in $[0,1]$, while orange means that only parameters existing in the training dataset are being used, \ie the marks of~\figref{fig:dimension_distribution}.
Our proposed version (purple) uses all possible dimensions sampled from a \ac{kde} of the training distribution, \ie the entire region in~\figref{fig:dimension_distribution}.

It can be observed that naturally \ac{mse} and \ac{fpd} increase when moving away from the distribution center of gravity (increasing~$\sigma$).
For random sampling, we see significantly worse performance in all three sampling categories compared to the other two methods.
For $1\sigma$ and $2\sigma$ the two other sampling methods perform almost equally well.
However, we see a significant performance improvement of our proposed sampling strategy in the most sparsely populated region where only 5\% of all training data lies.
This shows the advantage of the distribution-based training over a data-based training, especially for increasing~$\sigma$.

\subsection{Continuous Parameter Interpolation}

\input{figures/interpolation_chair_height}  

A decisive advantage of our method is that we can actively and directly influence the object dimensions.
\figref{fig:interpolation_chair_height} and \figref{fig:interpolation_table_width} show several examples of generated objects where the continuous parameter for the object size is changed.
For different latent vectors, our networks generates different shapes of good quality and diversity.
If now object sizes are changed, quality declines for the outermost samples (on the edge of the distribution).
However, our method still produces semantically meaningful objects in contrast to the baseline~B2.
For example, the overall shape of the office chair in the upper row of~\figref{fig:interpolation_chair_height} stays approximately the same while the backrest gets a slight tilt to the back to resemble an easy chair for lower heights, while B2 preserves the shape completely and simply compresses the object.
We want to stress that these figures are for demonstration purposes, the main application is rather to define the object size and then sample shapes (compare~\figref{fig:diversity}) and not to sample a shape and then modify the size, as shown here.

\input{figures/interpolation_table_width}  

\subsection{Out-of-Distribution Generation}

\figref{fig:eyecatcher}~shows a larger span of size conditioning, where only the samples enclosed in the dotted shape are conditioned on parameters sampled from within the distribution.
The dataset does not contain any chairs with sizes that are larger or smaller than the gray dots indicate (refer to~\figref{fig:dimension_distribution}).
The generated shapes for extreme low or high height or width do not necessarily resemble realistic objects, but the generator still maintains the approximate object and shape configuration.
All generated objects show smooth transitions between the configurations, even when out-of-distribution and do not collapse in shape.

\subsection{Model Properties}

\figref{fig:diversity}~shows five generated examples and five examples from the dataset that have the same dimensions.
The generated shapes are diverse and considerably distinct from the dataset samples.
This shows two things:
First, our method does not simply generate the same object for a given dimension when changing the latent vector (\emph{diversity}).
Second, it also does not learn a simple lookup by reproducing the samples from the dataset that lie close to the desired dimensions (\emph{novelty}).

As in related work, we demonstrate smooth representations within our model by interpolating between two latent vectors.
\figref{fig:latent_interpolation}~shows that we obtain smooth transitions from left to right for different object shapes.
This property was already included in the backbone network, but this experiment shows that it was not compromised by the introduction of our conditioning method.

\input{figures/diversity}  
\input{figures/latent_interpolation}  

\section{DISCUSSION}
\label{sec:discussion}

Influencing object dimensions is only one application out of many for explicit continuous conditioning.
\figref{fig:partseg}~demonstrates how our method can be used to influence the percentage of certain object parts relative to the entire shape.
To this end, the former dimension conditioning is now defined as a ratio of the body parts $\by \in \nR^2$ with $\by = (\text{wings}:\text{body}, \text{others}:\text{body}) \in [0, 1]^2$ for the class ``Airplane'' where $\sum \by = 1.0$.
The labels for the real data are obtained by using the point-wise part segmentation information from the dataset.
We count the number of points for each part and divide it with the number of points for the part ``body'' to obtain a ratio of points for each part.
Modifying the relation of object parts is more entangled with the overall appearance, therefore the shape changes considerably.
In contrast to the object dimensions it is not possible to trivially retrieve this parameter from the generated shape.
For the object dimensions a simple calculation suffices, while for the part manipulation a reliable part segmentation model would be needed, which requires annotations for training and can introduce additional errors.
Therefore, a naive regression loss for the conditioning vector or using a simple conditioning discriminator as in cGAN is not applicable.
This makes the discriminator regression vital.

\input{figures/interpolation_airplane_parts}  

Next steps for this work include the adaptation to real-world data.
Objects scanned with a laser are often only seen from a certain viewpoint in contrast to the objects in this work.
The viewpoint can be modeled by using additional continuous parameters that define the angle and distance from which the object is observed.
Regarding our proposed application for augmentation of autonomous driving LiDAR scenes, we can then define the size and position of a 3D bounding box for which the \ac{gan} then generates a custom-fit object.
A challenge is to adapt the backbone architecture to generate variable numbers of points, as common for real-world data.

\section{CONCLUSION}

This paper presented a novel \ac{gan} setup for 3D shape generation that uses continuous conditional parameters to actively influence the dimensions of generated shapes.
In extensive experiments we showed that our method can generate custom-fit objects that adhere the desired configuration while maintaining good generation quality and diversity.
We showed that our distribution label sampling is superior to sampling existing dataset parameters.
Further, we demonstrated generalization to out-of-distribution generation and gave a preview on potential other applications.
Future work includes the adaptation to real-world data.

\bibliography{bibliography/journals_long,bibliography/refs}


\clearpage
\appendix

\thispagestyle{empty}

\onecolumn \makesupplementtitle

\section{OVERVIEW}

This supplementary material covers details of the \ac{dnn} architectures, hyperparameters, evaluation, and additional results.
\secref{suppl:implementation}~gives all the training details of our proposed approach.
For the sake of completeness, \secref{suppl:analysis} discusses further experiments that are indicated in the main paper.
\secref{suppl:results}~presents further qualitative and quantitative results to complement the results section of the main paper.

\section{IMPLEMENTATION DETAILS}
\label{suppl:implementation}

\subsection{Architecture}

\tabref{tab:generator_implementation}~lists all layers, inputs, and operations of our \ac{dnn} architecture for the generator model.
We use the code\footnote{TreeGAN code: \url{https://github.com/seowok/TreeGAN}} from the original \emph{PyTorch} implementation of TreeGAN~\citep{Shu2019ICCV}.
Except for the input layers, our configuration is equal to the one of TreeGAN.
\tabref{tab:discriminator_implementation}~lists all layers, inputs, and operations of the discriminator \ac{dnn} architecture.
Here, the PointNet~\citep{Qi2017PointNet} contained in the TreeGAN code was used as a basis.
The split of the network for the auxiliary classifier mode is located directly after the PointNet feature extractor layers.
It is followed by two identical sequences of linear operations, where the adversarial head outputs a vector of size~$1$, while the regression head outputs a vector of size~$d$ of the continuous conditioning parameter.

\begin{table}
    \caption{
        \textbf{Generator Architecture}:
        Detailed network architecture and input format definition.
        The ID of each row is used to reference the output of the row.
        $\uparrow$ indicates that the layer directly above is an input.
        $d$ is the number of dimensions of the conditioning parameter.
        In case of the dimensions extent $d=3$, while for the influence of object part percentage $d=1$.
    }
    \label{tab:generator_implementation}
    \begin{tabularx}{\linewidth}{r p{3cm} X l}
        \toprule
        \textbf{ID} & \textbf{Inputs} & \textbf{Operation} & \textbf{Output Shape} \\
        \midrule
        1 & $\bz$ & Sample latent vector from $\bz~\sim~\mathcal{Z}=\mathcal{N}~(\mathbf{0},I)$ & $[96]$ \\
        2 & $\by_\text{cond}$ & Sample continuous parameter from $\by_\text{cond}~\sim~\operatorname{KDE}~(\by_\text{real})$ & $[d]$ \\
        \midrule
        \multicolumn{4}{c}{\textbf{Label Handling}} \\
        \midrule
        3 & 1 & Linear Layer & $[64]$ \\
        4 & 2 & Linear Layer & $[32]$ \\
        5 & 3, 4 & Concatenate & $[1 \times 96]$ \\
        \midrule
        \multicolumn{4}{c}{\textbf{Tree Graph Convolution (TreeGC) Network}} \\
        \midrule
        6 & $\uparrow$ & Tree Graph Convolution + LeakyReLU & $[1 \times 256]$ \\
        7 & $\uparrow$ & Branching & $[2 \times 256]$ \\
        8 & $\uparrow$ & Tree Graph Convolution + LeakyReLU & $[2 \times 256]$ \\
        9 & $\uparrow$ & Branching & $[4 \times 256]$ \\
        10 & $\uparrow$ & Tree Graph Convolution + LeakyReLU & $[4 \times 256]$ \\
        11 & $\uparrow$ & Branching & $[8 \times 256]$ \\
        12 & $\uparrow$ & Tree Graph Convolution + LeakyReLU & $[8 \times 128]$ \\
        13 & $\uparrow$ & Branching & $[16 \times 128]$ \\
        14 & $\uparrow$ & Tree Graph Convolution + LeakyReLU & $[16 \times 128]$ \\
        15 & $\uparrow$ & Branching & $[32 \times 128]$ \\
        16 & $\uparrow$ & Tree Graph Convolution + LeakyReLU & $[32 \times 128]$ \\
        17 & $\uparrow$ & Branching & $[2048 \times 128]$ \\
        18 & $\uparrow$ & Tree Graph Convolution & $[2048 \times 3]$ \\
        \bottomrule
    \end{tabularx}
\end{table}
\begin{table}
    \caption{
        \textbf{Discriminator Architecture}:
        Detailed network architecture and input format definition.
        The ID of each row is used to reference the output of the row.
        $\uparrow$ indicates that the layer directly above is an input.
        $d$ is the number of dimensions of the conditioning parameter.
        In case of the dimensions extend $d=3$, while for the influence of object part percentage $d=1$.
    }
    \label{tab:discriminator_implementation}
    \begin{tabularx}{\linewidth}{r l X l X}
        \toprule
        \textbf{ID} & \textbf{Inputs} & \textbf{Operation} & \textbf{Output Shape} & \textbf{Description} \\
        \midrule
        1 & point cloud~$\bx$ & $x, y, z$ & $[2048 \times 3]$ & Input point cloud $\bx=\bx_\text{real}$ for real data and $\bx=\bx_\text{gen}$ for generated data. \\
        \midrule
        \multicolumn{5}{c}{\textbf{PointNet Feature Extractor}} \\
        \midrule
        2 & $\uparrow$ & Conv1D + LeakyReLU & $[2048 \times 64]$ & Kernel size $1\times1$, stride 1 \\
        3 & $\uparrow$ & Conv1D + LeakyReLU & $[2048 \times 128]$ & Kernel size $1\times1$, stride 1 \\
        4 & $\uparrow$ & Conv1D + LeakyReLU & $[2048 \times 256]$ & Kernel size $1\times1$, stride 1 \\
        5 & $\uparrow$ & Conv1D + LeakyReLU & $[2048 \times 512]$ & Kernel size $1\times1$, stride 1 \\
        6 & $\uparrow$ & Conv1D + LeakyReLU & $[2048 \times 1024]$ & Kernel size $1\times1$, stride 1 \\
        7 & $\uparrow$ & MaxPool & $[1024]$ & Global features \\
        \midrule
        \multicolumn{5}{c}{\textbf{Adversarial Output Head}} \\
        \midrule
        8 & $\uparrow$ & Linear Layer & $[1024]$ & \\
        9 & $\uparrow$ & Linear Layer & $[512]$ & \\
        10 & $\uparrow$ & Linear Layer & $[512]$ & \\
        11 & $\uparrow$ & Linear Layer & $[1]$ & Output vector $D(\bx)$\\
        \midrule
        \multicolumn{5}{c}{\textbf{Regression Output Head}} \\
        \midrule
        12 & 7 & Linear Layer & $[1024]$ & \\
        13 & $\uparrow$ & Linear Layer & $[512]$ & \\
        14 & $\uparrow$ & Linear Layer & $[512]$ & \\
        15 & $\uparrow$ & Linear Layer & $[d]$ & Output vector $\hat{\by}(\bx)$ \\
        \bottomrule
    \end{tabularx}
\end{table}

\subsection{Training}

Two Adam optimizers are used for optimization, one for the parameters of the generator and one for the discriminator.
For both, the learning rate is set to $1e^{-4}$.
Additionally, the two weighting factors $\lambda_\text{adv}$ and $\lambda_\text{reg}$ for the losses are optimized.
In the main paper, we formulate the losses for the generator and discriminator as
\begin{equation*}
    \mathcal{L} = \lambda_\text{adv} \cdot \mathcal{L}_\text{adv} + \lambda_\text{reg} \cdot \mathcal{L}_\text{reg}
\end{equation*}
with the adversarial loss $\mathcal{L}_\text{adv}$ and the regression loss $\mathcal{L}_\text{reg}$.
In order to avoid simply learning weighting factors of zero and to ensure stable training convergence at the same time, the loss is implemented as
\begin{equation*}
    \mathcal{L} = \mathcal{L}_\text{adv} \cdot e^{v_\text{adv}} + v_\text{adv} + \mathcal{L}_\text{reg} \cdot e^{v_\text{reg}} + v_\text{reg}
\end{equation*}
with $v_\text{adv}$ and $v_\text{reg}$ being the trainable variables.
Both variables are initialized to $v=0$ at the beginning of the training, such that both loss parts are equally weighted.

\section{ADDITIONAL ANALYSIS}
\label{suppl:analysis}

\subsection{Loss Variations}

Both the generator and the discriminator loss consist of an adversarial part and a regression part.
The generator regression loss computes the error between the requested parameter $\by_\text{cond}$ and the corresponding discriminator prediction, while the discriminator regression is defined as the error between the parameter of the real data $\by_\text{real}$ and its corresponding prediction $\hat{\by}_\text{real}$, such that
\begin{align*}
    \mathcal{L}_{G,\text{reg}} &= \mathcal{L}_\text{reg} \left( \by_\text{cond}, \hat{\by}_\text{gen} \right)
    \quad \text{and} \\
    \mathcal{L}_{D,\text{reg}} &= \mathcal{L}_\text{reg} \left( \by_\text{real}, \hat{\by}_\text{real} \right) .
\end{align*}

It is notable that $\mathcal{L}_{D,\text{reg}}$ is only computed for the real samples and not for the generated samples and that $\mathcal{L}_{G,\text{reg}}$ uses the discriminator prediction~$\hat{\by}_\text{gen}$ instead of the configuration of the actually generated object~$\by_\text{gen}$.
These design choices can be attributed to the fact that in many cases the actual parameter~$\by_\text{gen}$ of the generated point cloud~$\bx_\text{gen}$ is unknown, \ie cannot be trivially retrieved from $\bx_\text{gen}$.
Since our application of influencing object dimensions offers the possibility to simply compute $\by_\text{gen}$ from $\bx_\text{gen}$ with $\by_\text{gen}=\lVert \max(\bx_\text{gen}) - \min(\bx_\text{gen}) \rVert$, we additionally investigated loss variations that exploit this property.

First, we define the generator and discriminator regression losses as
\begin{align*}
    \mathcal{L}_{G,\text{reg}} &= \mathcal{L}_\text{reg} \left( \by_\text{cond}, \by_\text{gen} \right)
    \quad \text{and} \\
    \mathcal{L}_{D,\text{reg}} &= \frac{1}{2} \Big[
        \mathcal{L}_\text{reg} \left( \by_\text{real}, \hat{\by}_\text{real} \right) +
        \mathcal{L}_\text{reg} \left( \by_\text{gen}, \hat{\by}_\text{gen} \right)
    \Big],
\end{align*}
respectively.
We found that this leads to slightly better training convergence, but no significant outperformance in the final model in terms of \ac{fpd} or \ac{mse}.
Therefore, we conclude that the proposed losses of the main paper are a good mechanism to train the model when $\by_\text{gen}$ is unknown, as for most applications.

Second, we investigated another loss configuration.
As for the discriminator, our proposed method skips the generated part of the loss entirely.
However, it is also possible to formulate the losses as follows
\begin{align*}
    \mathcal{L}_{G,\text{reg}} &= \mathcal{L}_\text{reg} \left( \by_\text{cond}, \hat{\by}_\text{gen} \right)
    \quad \text{and} \\
    \mathcal{L}_{D,\text{reg}} &= \frac{1}{2} \Big[
        \mathcal{L}_\text{reg} \left( \by_\text{real}, \hat{\by}_\text{real} \right) +
        \mathcal{L}_\text{reg} \left( \by_\text{cond}, \hat{\by}_\text{gen} \right)
    \Big],
\end{align*}
where $\by_\text{gen}$ from above is replaced with $\hat{\by}_\text{real}$ in the generator and $\by_\text{cond}$ in the discriminator.
However, we found that this leads to unstable training and results in a significantly worse final model.
We attribute this to false feedback for the discriminator, especially in the beginning of the training.
At that time, the generator is not yet well enough trained to output shapes that are close to the requested parameters ($\by_\text{gen} \neq \hat{\by}_\text{gen}$).
Therefore, we found that it is best to not include the generated path for the discriminator regression at all.

\subsection{Other Conditioning Concepts}

Additionally to our proposed method we also investigate other configurations for the continuous conditioning of point cloud generation.
These method either did not yield promising results or are limited in applicability, therefore they are not included in the main paper.
However, for the sake of completeness and reproducibility, we include all relevant implementation details here and state the overall results of our experiments.

\subsubsection{cGAN with Continuous Parameters}

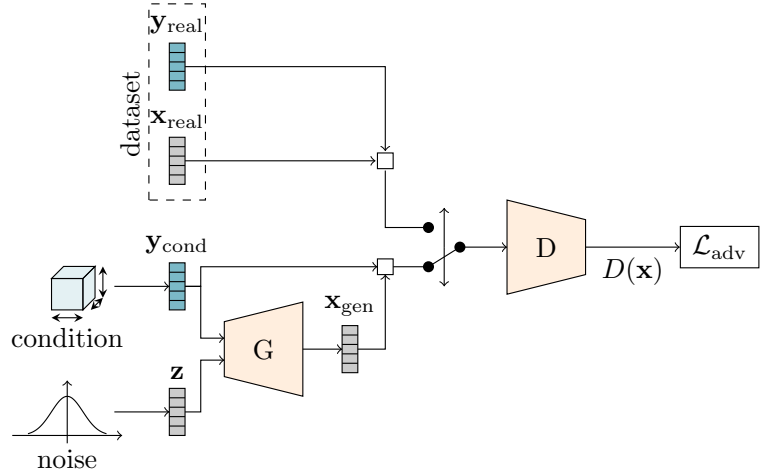
\begin{figure}
	\centering
	\resizebox{0.6\linewidth}{!}{\begin{tikzpicture}
\pgfdeclarelayer{bg}    
\pgfsetlayers{bg,main}  
\definecolor{label}{RGB}{123, 185, 199}
\definecolor{network}{RGB}{255, 238, 222}
\tikzset{
    vecinp/.pic = {
        \path[pic actions, draw=black] (-0.1,-0.3) rectangle (0.1,0.3);
        \draw (-0.1, -0.18) -- (0.1, -0.18);
        \draw (-0.1, -0.06) -- (0.1, -0.06);
        \draw (-0.1, 0.06) -- (0.1, 0.06);
        \draw (-0.1, 0.18) -- (0.1, 0.18);

        \node[draw=none, text=black, opacity=1.0, anchor=south] at (0,0.3) {\tikzpictext};
    },
    connsq/.pic = {
        \def\w{0.2};
        \path[draw=black] (-\w/2,-\w/2) rectangle +(\w,\w);
    }
}

\draw[white] (0.8,-1.2) rectangle (10.5,5.0);

\coordinate (preal) at (6.2,2.0);
\coordinate (pfake) at (6.2,1.5);
\coordinate (pdis) at (6.6,1.75);

\coordinate (Generator) at (4.1,0.45);
\coordinate (Discriminator) at ($(pdis)+(1.1,0)$);

\coordinate (xgen) at ($(Generator)+(1.1, 0)$);
\coordinate (ygen) at ($(xgen)+(3.4, -0.4)$);

\coordinate (calc) at ($(preal)-(3, 0)$);
\coordinate (y) at ($(calc)+(0.8,0)$);

\coordinate (Dr) at ($(preal)-(0.55,0)$);
\coordinate (Df) at ($(pfake)-(0.55,0)$);

\fill (preal) circle (2pt);
\fill (pfake) circle (2pt);
\fill (pdis) circle (2pt);

\draw (pfake) -- (pdis);
\draw[->] ($($(preal)!0.5!(pfake)$)!0.5!(pdis)$) -- +(0,-0.5);
\draw[->] ($($(preal)!0.5!(pfake)$)!0.5!(pdis)$) -- +(0,0.5);
\draw[->] (pdis) -- ($(Discriminator)-(0.5,0)$);

\begin{scope}[shift={(1.4,0.15)}]
    \begin{scope}[shift={(0,0.8)}]
        \begin{scope}[shift={(0.2,0)}]
            \draw[fill=label, fill opacity=0.2] (-0.2, 0.0, 0.0) -- +(0.4, 0, 0) -- +(0.4, 0.4, 0) -- +(0, 0.4, 0) -- cycle;
            \draw[fill=label, fill opacity=0.2] (0.2, 0.0, 0.0) -- +(0, 0, -0.4) -- +(0, 0.4, -0.4) -- +(0, 0.4, 0) -- cycle;
            \draw[fill=label, fill opacity=0.2] (0.2, 0.4, 0.0) -- +(0, 0, -0.4) -- +(-0.4, 0, -0.4) -- +(-0.4, 0, 0) -- cycle;

            \draw[stealth-stealth] (-0.2, -0.1, 0.0) -- +(0.4, 0, 0);
            \draw[stealth-stealth] (0.3, 0.0, 0.0) -- +(0, 0, -0.4);
            \draw[stealth-stealth] (0.3, 0.0, -0.4) -- +(0, 0.4, 0);

            \node[draw=none, anchor=north] at (0,-0.1) {condition};
        \end{scope}

        \coordinate (yhat) at (1.6, 0.3);
        \draw[->] (0.8,0.3) -- ($(yhat)-(0.1,0)$);
        \pic[fill = label, pic text = {$\by_\text{cond}$}] at (yhat) {vecinp};
    \end{scope}

    \begin{scope}[shift={(0,-0.8)}]
        \begin{scope}[scale=0.5,shift={(0.4,0)}]
            \draw[domain=-1:1, smooth, variable=\x, black]  plot ({\x}, {1/(0.4*sqrt(2*pi))*exp(-((\x-0.0)^2)/(2*0.4^2))});
            \draw[->] (-1.4,0) -- (1.4,0);
            \draw[->] (0,-0.2) -- (0,1.4);
            \node[draw=none, anchor=north] at (0,-0.1) {noise};
        \end{scope}

        \coordinate (z) at (1.6,0.3);
        \draw[->] (0.8,0.3) -- ($(z)-(0.1,0)$);
        \pic[fill = white!80!black, pic text = {$\bz$}] at (z) {vecinp};
    \end{scope}

\end{scope}

\draw[->] ($(yhat)+(0.1, 0)$) -| +(0.2, 0) |- ($(Generator)-(0.5,-0.14)$);
\draw[->] ($(z)+(0.1, 0)$) -| +(0.2, 0) |- ($(Generator)-(0.5,0.14)$);

\begin{scope}[shift={(Generator)}]
    \fill[network, draw=black] (-0.5,-0.35) -- (0.5,-0.6) -- (0.5,0.6) -- (-0.5,0.35) -- cycle;
    \node[draw=none, anchor=center] at (0,0) {G};
\end{scope}

\pic [fill = white!80!black, pic text = {$\bx_\text{gen}$}] at (xgen) {vecinp};

\draw[->] ($(Generator)+(0.5,0)$) -- ($(xgen)-(0.1,0)$);  

\draw[fill=white, draw=black] ($(Df)-(0.1,0.1)$) rectangle +(0.2,0.2);
\draw ($(Df)+(0.1,0)$) -- (pfake);
\draw[->] ($(xgen)+(0.1,0)$) -| ($(Df)-(0,0.1)$);  
\draw[->] ($(yhat)+(0.1, 0)$) -| +(0.2, 0) |- ($(Df)-(0.1,0)$); 

\coordinate (x) at ($(yhat)+(0,1.6)$);
\coordinate (y) at ($(x)+(0,1.2)$);

\pic [fill = white!80!black, pic text = {$\bx_\text{real}$}] at (x) {vecinp};
\pic [fill = label, pic text = {$\by_\text{real}$}] at (y) {vecinp};
\draw[dashed] ($(x)-(0.36,0.5)$) rectangle ($(y)+(0.36,0.8)$);
\node[rotate=90] at ($(x)!0.5!(y)-(0.6,0)$) {dataset};

\path ($(x)+(0.1,0)$) -| ($(preal)-(0.55,0)$) node[pos=0.5] (s) {} |- (preal);
\draw[fill=white, draw=black] ($(s)-(0.1,0.1)$) rectangle +(0.2,0.2);

\draw[->] ($(y)+(0.1,0)$) -| (s);
\draw[->] ($(x)+(0.1,0)$) -- (s);
\draw (s) |- (preal);

\begin{scope}[shift={(Discriminator)}]
    \fill[network, draw=black] (-0.5,-0.6) -- (0.5,-0.35) -- (0.5,0.35) -- (-0.5,0.6) -- cycle;
    \node[draw=none, anchor=center] at (0,0) {D};
\end{scope}

\draw[->] ($(Discriminator)+(0.5,0)$) -- +(1.2, 0) node[anchor=north, pos=0.5] {$D(\bx)$}  node[anchor=west, draw=black, minimum width=1cm] {$\mathcal{L}_\text{adv}$};

\end{tikzpicture}}
	\caption{
        \textbf{cGAN with continuous parameters}:
        The generator~$G$ generates a point cloud~$\bx_\text{gen}$ from a random vector~$\bz$ and a continuous parameter~$\by$.
		The discriminator either receives a set of point cloud and parameter from the real~$\{\by_\text{real},\bx_\text{real}\}$ or the generated distribution~$\{\by_\text{cond},\bx_\text{gen}\}$.
		It then outputs an estimate whether the set is real or generated with which the adversarial loss~$\mathcal{L}_\text{adv}$ is computed.
    }
	\label{fig:cgan}
\end{figure}

We adapted the concepts of cGAN~\figref{fig:cgan} and CcGAN~\citep{Ding2021ICLR} to work with TreeGAN~\citep{Shu2019ICCV} as the backbone network.
In contrast to our proposed approach, we call this as an implicit conditioning scheme, since there is no extrinsic excitation that forces the model to use the conditioning input explicitly.
The discriminator receives both the point cloud and the conditioning parameter as an input.
The loss function of the generator~$G$ is defined as
\begin{equation*}
    \mathcal{L}_G =
    - \nE_{\bz \sim \mathcal{Z}} \left[ D\left( \by_\text{cond}, G(\by_\text{cond}, \bz) \right) \right]
\end{equation*}
where $\mathcal{Z}$ represents the latent code distribution which follows a Normal distribution, such that $z \in \mathcal{N}(0,1)$.
The loss function of the discriminator is defined as
\begin{align*}
    \mathcal{L}_D =
    &\nE_{\bz \sim \mathcal{Z}} \left[ D\left( \by_\text{cond}, G(\by_\text{cond}, \bz) \right) \right] \\
    &- \nE_{\bx \sim \mathcal{R}} \left[ D\left( \by_\text{real}, \bx \right) \right]
    + \mathcal{L}_\text{gp}
\end{align*}
with the gradient penalty~$\mathcal{L}_\text{gp}$ as defined in the main paper.

\begin{figure}
	\centering

	\begin{subfigure}[t]{0.24\linewidth}
		\centering
		\resizebox{\linewidth}{!}{\begin{tikzpicture}
\definecolor{label}{RGB}{123, 185, 199}
\definecolor{network}{RGB}{255, 238, 222}
\definecolor{tree}{RGB}{125, 173, 250}
\tikzset{
    vecinp/.pic = {
        \path[pic actions, draw=black] (-0.1,-0.3) rectangle (0.1,0.3);
        \draw (-0.1, -0.18) -- (0.1, -0.18);
        \draw (-0.1, -0.06) -- (0.1, -0.06);
        \draw (-0.1, 0.06) -- (0.1, 0.06);
        \draw (-0.1, 0.18) -- (0.1, 0.18);

        \node[draw=none, text=black, opacity=1.0, anchor=south] at (0,0.3) {\tikzpictext};
    },
}

\draw[white] (-1.2, -2) rectangle (4.6, 2);

\fill[network, draw=black] (0,-1.2) -- (3.2,-1.8) -- (3.2,1.8) -- (0,1.2) -- cycle;

\coordinate (y) at (-0.5, 0.55);
\coordinate (z) at (-0.5, -0.55);

\pic[fill = label, pic text = {$\by_\text{cond}$}] at (y) {vecinp};
\pic[fill = white!80!black, pic text = {$\bz$}] at (z) {vecinp};

\node[draw=black, fill=white, rotate=90] (concat) at (0.6, 0) {concat};

\pic[fill = label] at (1.3, 0.3) {vecinp};
\pic[fill = white!80!black] at (1.3, -0.3) {vecinp};

\fill[tree, draw=black] (2, -1.2) rectangle +(1, 2.4) node[text=black, pos=.5, text width=0.8cm, align=center] (tree) {Tree GCN};

\pic[fill = white!80!black, pic text = {$\bx_\text{gen}$}] at (3.7, 0) {vecinp};

\draw[->] ($(y)+(0.1,0)$) -| ($(concat.north)-(0.2, -0.2)$) |- ($(concat.north)-(0, -0.2)$);
\draw[->] ($(z)+(0.1,0)$) -| ($(concat.north)-(0.2, 0.2)$) |- ($(concat.north)-(0, 0.2)$);

\draw[->] (concat.south) -- (1.2, 0);
\draw[->] (1.4, 0) -- (tree.west);
\draw[->] (tree.east) -- (3.6, 0);

\end{tikzpicture}}
		\caption{\label{fig:vanilla_generator}Vanilla Generator}
	\end{subfigure}%
	\hspace{0.5em}%
	\begin{subfigure}[t]{0.24\linewidth}
		\centering
		\resizebox{\linewidth}{!}{\begin{tikzpicture}[
    plus/.style={path picture={
        \draw[black] (path picture bounding box.south) -- (path picture bounding box.north) (path picture bounding box.west) -- (path picture bounding box.east);
    }},
    times/.style={path picture={
        \draw[black] (path picture bounding box.south west) -- (path picture bounding box.north east) (path picture bounding box.north west) -- (path picture bounding box.south east);
    }}
]
\definecolor{label}{RGB}{123, 185, 199}
\definecolor{network}{RGB}{255, 238, 222}
\definecolor{pointnet}{RGB}{125, 173, 250}
\definecolor{fc}{RGB}{116, 219, 173}
\tikzset{
    vecinp/.pic = {
        \path[pic actions, draw=black] (-0.1,-0.3) rectangle (0.1,0.3);
        \draw (-0.1, -0.18) -- (0.1, -0.18);
        \draw (-0.1, -0.06) -- (0.1, -0.06);
        \draw (-0.1, 0.06) -- (0.1, 0.06);
        \draw (-0.1, 0.18) -- (0.1, 0.18);

        \node[draw=none, text=black, opacity=1.0, anchor=south] at (0,0.3) {\tikzpictext};
    },
}

\draw[white] (-1.2, -2) rectangle (4.6, 2);

\fill[network, draw=black] (0,-1.8) -- (3.2,-1.2) -- (3.2,1.2) -- (0,1.8) -- cycle;

\coordinate (net) at (0.7, 0);

\fill[pointnet, draw=black] ($(net)-(0.5,1.4)$) rectangle +(1, 2) node[text=black, pos=.5, text width=0.8cm, align=center] (pointnet) {Point Net};

\coordinate (x) at ($(pointnet)-(1.2,0)$);
\coordinate (y) at ($(x)+(0,1.45)$);

\pic[fill = label, pic text = {$\by$}] at (y) {vecinp};
\pic[fill = white!80!black, pic text = {$\bx$}] at (x) {vecinp};

\coordinate (vec) at ($(pointnet)+(0.8,0)$);
\pic[fill = white!50!black] at (vec) {vecinp};

\node[draw=black, fill=white, rotate=90] (concat) at (2.1, 0) {concat};

\fill[fc, draw=black] (2.6, -0.5) rectangle +(0.5, 1) node[text=black, pos=.5, rotate=90, text width=0.8cm, align=center] (linear) {FC};

\draw[->] ($(y)+(0.1,0)$) -| ($(concat.north)+(-0.15,0.2)$) |- ($(concat.north)+(0,0.2)$);

\draw[->] ($(x)+(0.1,0)$) -- (pointnet.west);
\draw[->] (pointnet.east) -- ($(vec)-(0.1,0)$);
\draw[->] ($(vec)+(0.1,0)$) -| ($(concat.north)-(0.15,0.2)$) |- ($(concat.north)-(0,0.2)$);

\draw[->] (concat.south) -- (linear.north);
\draw[->] (linear.south) -- +(0.4, 0) node[anchor=west] {$D(\bx)$};

\end{tikzpicture}}
		\caption{\label{fig:vanilla_discriminator}Vanilla Discriminator}
	\end{subfigure}
	\hspace{0.5em}%
	\begin{subfigure}[t]{0.24\linewidth}
		\centering
		\resizebox{\linewidth}{!}{\begin{tikzpicture}[
    plus/.style={path picture={
        \draw[black] (path picture bounding box.south) -- (path picture bounding box.north) (path picture bounding box.west) -- (path picture bounding box.east);
    }}
]
\definecolor{label}{RGB}{123, 185, 199}
\definecolor{network}{RGB}{255, 238, 222}
\definecolor{tree}{RGB}{125, 173, 250}
\definecolor{fc}{RGB}{116, 219, 173}
\tikzset{
    vecinp/.pic = {
        \path[pic actions, draw=black] (-0.1,-0.3) rectangle (0.1,0.3);
        \draw (-0.1, -0.18) -- (0.1, -0.18);
        \draw (-0.1, -0.06) -- (0.1, -0.06);
        \draw (-0.1, 0.06) -- (0.1, 0.06);
        \draw (-0.1, 0.18) -- (0.1, 0.18);

        \node[draw=none, text=black, opacity=1.0, anchor=south] at (0,0.3) {\tikzpictext};
    },
}

\draw[white] (-1.2, -2) rectangle (4.6, 2);

\fill[network, draw=black] (0,-1.2) -- (3.2,-1.8) -- (3.2,1.8) -- (0,1.2) -- cycle;

\coordinate (y) at (-0.5, 0.55);
\coordinate (z) at (-0.5, -0.55);
\coordinate (fcy) at (0.5, 0.55);
\coordinate (fcz) at (0.5, -0.55);
\coordinate (vin) at (1.6, 0);

\pic[fill = label, pic text = {$\by_\text{cond}$}] at (y) {vecinp};
\pic[fill = white!80!black, pic text = {$\bz$}] at (z) {vecinp};

\fill[fc, draw=black] ($(fcy)-(0.25,0.5)$) rectangle +(0.5, 1) node[text=black, pos=.5, rotate=90] (lineary) {FC};
\fill[fc, draw=black] ($(fcz)-(0.25,0.5)$) rectangle +(0.5, 1) node[text=black, pos=.5, rotate=90] (linearz) {FC};
\node[draw, fill=white, circle, plus, minimum width=0.1 cm] (P) at (1.1,0) {};
\pic[fill = white!50!black] at (vin) {vecinp};

\fill[tree, draw=black] (2, -1.2) rectangle +(1, 2.4) node[text=black, pos=.5, text width=0.8cm, align=center] (tree) {Tree GCN};

\pic[fill = white!80!black, pic text = {$\bx_\text{gen}$}] at (3.7, 0) {vecinp};

\draw[->] ($(y)+(0.1,0)$) -- (lineary.north);
\draw[->] (lineary) -| (P);

\draw[->] ($(z)+(0.1,0)$) -- (linearz.north);
\draw[->] (linearz) -| (P);

\draw[->] (P) -- ($(vin)-(0.1,0)$);
\draw[->] ($(vin)+(0.1,0)$) -- (tree.west);
\draw[->] (tree.east) -- (3.6, 0);

\end{tikzpicture}}
		\caption{\label{fig:projection_generator}Projection Generator}
	\end{subfigure}%
	\hspace{0.5em}%
	\begin{subfigure}[t]{0.24\linewidth}
		\centering
		\resizebox{\linewidth}{!}{\begin{tikzpicture}[
    plus/.style={path picture={
        \draw[black] (path picture bounding box.south) -- (path picture bounding box.north) (path picture bounding box.west) -- (path picture bounding box.east);
    }},
    times/.style={path picture={
        \draw[black] (path picture bounding box.south west) -- (path picture bounding box.north east) (path picture bounding box.north west) -- (path picture bounding box.south east);
    }}
]
\definecolor{label}{RGB}{123, 185, 199}
\definecolor{network}{RGB}{255, 238, 222}
\definecolor{pointnet}{RGB}{125, 173, 250}
\definecolor{fc}{RGB}{116, 219, 173}
\tikzset{
    vecinp/.pic = {
        \path[pic actions, draw=black] (-0.1,-0.3) rectangle (0.1,0.3);
        \draw (-0.1, -0.18) -- (0.1, -0.18);
        \draw (-0.1, -0.06) -- (0.1, -0.06);
        \draw (-0.1, 0.06) -- (0.1, 0.06);
        \draw (-0.1, 0.18) -- (0.1, 0.18);

        \node[draw=none, text=black, opacity=1.0, anchor=south] at (0,0.3) {\tikzpictext};
    },
}

\draw[white] (-1.2, -2) rectangle (4.6, 2);

\fill[network, draw=black] (0,-1.8) -- (3.2,-1.2) -- (3.2,1.2) -- (0,1.8) -- cycle;

\coordinate (net) at (0.7, 0);

\fill[pointnet, draw=black] ($(net)-(0.5,1.4)$) rectangle +(1, 2) node[text=black, pos=.5, text width=0.8cm, align=center] (pointnet) {Point Net};
\fill[fc, draw=black] ($(net)+(-0.5, 0.8)$) rectangle +(1, 0.5) node[text=black, pos=.5, text width=0.8cm, align=center] (linear1) {FC};

\coordinate (y) at ($(linear1)-(1.2,0)$);
\coordinate (x) at ($(pointnet)-(1.2,0)$);

\pic[fill = label, pic text = {$\by$}] at (y) {vecinp};
\pic[fill = white!80!black, pic text = {$\bx$}] at (x) {vecinp};

\coordinate (vec) at ($(pointnet)+(0.8,0)$);
\pic[fill = white!50!black] at (vec) {vecinp};

\fill[fc, draw=black] ($(vec)+(0.4,-0.5)$) rectangle +(0.5, 1) node[text=black, pos=.5, rotate=90, text width=0.8cm, align=center] (linear2) {FC};

\node[draw, fill=white, circle, times, minimum width=0.1 cm] (T) at ($(linear1)+(1,0)$) {};
\node[draw, fill=white, circle, plus, minimum width=0.1 cm] (P) at (2.9, 0) {};

\draw[->] ($(y)+(0.1,0)$) -- (linear1.west);
\draw[->] (linear1.east) -- (T.west);
\draw[->] (T.east) -| (P.north);

\draw[->] ($(x)+(0.1,0)$) -- (pointnet.west);
\draw[->] (pointnet.east) -- ($(vec)-(0.1,0)$);
\draw[->] ($(vec)+(0.1,0)$) -- (linear2.north);
\draw[->] ($(vec)+(0.1,0)$) -| (T.south);
\draw[->] (linear2.south) -| (P.south);
\draw[->] (P.east) -- +(0.4, 0) node[anchor=west] {$D(\bx)$};

\end{tikzpicture}}
		\caption{\label{fig:projection_discriminator}Projection Discriminator}
	\end{subfigure}

	\caption{
		\textbf{Parameter handling details}:
		Generator and discriminator input handling for Vanilla and Projection cGAN.
	}

	\label{fig:cgan_details}
\end{figure}
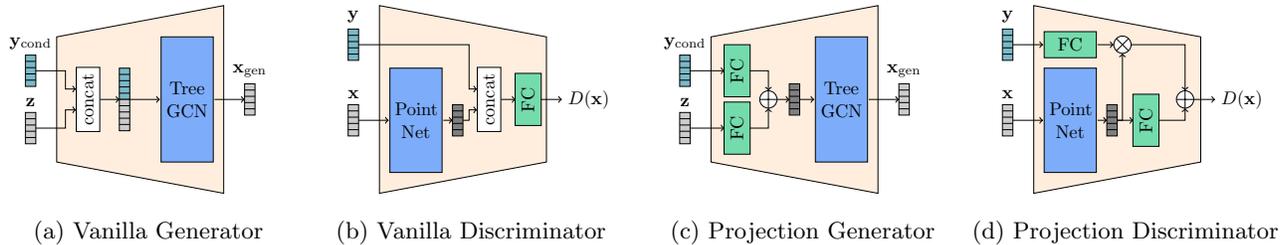

For the first variant, we use the traditional label incorporation introduced by cGAN~\citep{Mirza2014}.
We refer to this as the Vanilla cGAN.
The generator details are depicted in~\figref{fig:vanilla_generator} and the discriminator details are shown in~\figref{fig:vanilla_discriminator}.
The generator label incorporation simply concatenates the latent and the label vector prior to feeding it to the graph convolution network~(GCN).
In the discriminator, the point cloud is processed by PointNet which outputs a feature vector to which the label vector is concatenated before being processed by a final set of linear layers.

For the second variant, we use the label input configuration proposed by CcGAN~\citep{Ding2021ICLR} to handle continuous parameters.
Their approach is inspired by the method of label projection~\citep{Miyato2018ICLR}, which is why we refer to this variant as Projection cGAN.
We used HVDL+NLI and conducted a coarse hyper-parameter search in steps of magnitudes (\eg $10^{..1,2,3,..}$ to cover a wide range of options.
Details for the generator and discriminator are shown in \figref{fig:projection_generator} and \figref{fig:projection_discriminator}, respectively.
In contrast to Vanilla cGAN, both the latent and the label vectors are passed through a linear layer first, after which both are added together element-wise.
For the discriminator, the label vector is fed through a linear layer after which the inner product with the features from PointNet is calculated.
The features are fed through the final set of linear layers after which the result is added to the result of the inner product.

As mentioned in the main paper, both variants did not achieve satisfying results.
The Vanilla variant ignored the conditioning entirely which led to very high \ac{mse} values (about four magnitudes higher than our proposed method).
The performance in terms of \ac{fpd} is close to the backbone.
For the Projection variant, we observed very unstable training courses that often led to a collapse of the training with the model performing significantly worse in all metrics compared to all other models.

\subsubsection{cGAN with Regression}

\begin{figure}
	\centering
	\resizebox{0.6\linewidth}{!}{\begin{tikzpicture}
\pgfdeclarelayer{bg}    
\pgfsetlayers{bg,main}  
\definecolor{label}{RGB}{123, 185, 199}
\definecolor{network}{RGB}{255, 238, 222}
\tikzset{
    vecinp/.pic = {
        \path[pic actions, draw=black] (-0.1,-0.3) rectangle (0.1,0.3);
        \draw (-0.1, -0.18) -- (0.1, -0.18);
        \draw (-0.1, -0.06) -- (0.1, -0.06);
        \draw (-0.1, 0.06) -- (0.1, 0.06);
        \draw (-0.1, 0.18) -- (0.1, 0.18);

        \node[draw=none, text=black, opacity=1.0, anchor=south] at (0,0.3) {\tikzpictext};
    },
    connsq/.pic = {
        \def\w{0.2};
        \path[draw=black] (-\w/2,-\w/2) rectangle +(\w,\w);
    }
}

\draw[white] (0.8,-1.2) rectangle (10.5,5.0);

\coordinate (preal) at (6.2,2.0);
\coordinate (pfake) at (6.2,1.5);
\coordinate (pdis) at (6.6,1.75);

\coordinate (Generator) at (4.1,0.45);
\coordinate (Discriminator) at ($(pdis)+(1.1,0)$);

\coordinate (xgen) at ($(Generator)+(1.1, 0)$);
\coordinate (ygen) at ($(xgen)+(3.4, -0.4)$);

\fill (preal) circle (2pt);
\fill (pfake) circle (2pt);
\fill (pdis) circle (2pt);

\draw (pfake) -- (pdis);
\draw[->] ($($(preal)!0.5!(pfake)$)!0.5!(pdis)$) -- +(0,-0.5);
\draw[->] ($($(preal)!0.5!(pfake)$)!0.5!(pdis)$) -- +(0,0.5);
\draw[->] (pdis) -- ($(Discriminator)-(0.5,0)$) node[anchor=south, pos=0.5] {$\bx$};

\begin{scope}[shift={(1.4,0.15)}]
    \begin{scope}[shift={(0,0.8)}]
        \begin{scope}[shift={(0.2,0)}]
            \draw[fill=label, fill opacity=0.2] (-0.2, 0.0, 0.0) -- +(0.4, 0, 0) -- +(0.4, 0.4, 0) -- +(0, 0.4, 0) -- cycle;
            \draw[fill=label, fill opacity=0.2] (0.2, 0.0, 0.0) -- +(0, 0, -0.4) -- +(0, 0.4, -0.4) -- +(0, 0.4, 0) -- cycle;
            \draw[fill=label, fill opacity=0.2] (0.2, 0.4, 0.0) -- +(0, 0, -0.4) -- +(-0.4, 0, -0.4) -- +(-0.4, 0, 0) -- cycle;

            \draw[stealth-stealth] (-0.2, -0.1, 0.0) -- +(0.4, 0, 0);
            \draw[stealth-stealth] (0.3, 0.0, 0.0) -- +(0, 0, -0.4);
            \draw[stealth-stealth] (0.3, 0.0, -0.4) -- +(0, 0.4, 0);

            \node[draw=none, anchor=north] at (0,-0.1) {condition};
        \end{scope}

        \coordinate (yhat) at (1.6, 0.3);
        \draw[->] (0.8,0.3) -- ($(yhat)-(0.1,0)$);
        \pic[fill = label, pic text = {$\by_\text{cond}$}] at (yhat) {vecinp};
    \end{scope}

    \begin{scope}[shift={(0,-0.8)}]
        \begin{scope}[scale=0.5,shift={(0.4,0)}]
            \draw[domain=-1:1, smooth, variable=\x, black]  plot ({\x}, {1/(0.4*sqrt(2*pi))*exp(-((\x-0.0)^2)/(2*0.4^2))});
            \draw[->] (-1.4,0) -- (1.4,0);
            \draw[->] (0,-0.2) -- (0,1.4);
            \node[draw=none, anchor=north] at (0,-0.1) {noise};
        \end{scope}

        \coordinate (z) at (1.6,0.3);
        \draw[->] (0.8,0.3) -- ($(z)-(0.1,0)$);
        \pic[fill = white!80!black, pic text = {$\bz$}] at (z) {vecinp};
    \end{scope}

\end{scope}

\draw[->] ($(yhat)+(0.1, 0)$) -| +(0.2, 0) |- ($(Generator)-(0.5,-0.14)$);
\draw[->] ($(z)+(0.1, 0)$) -| +(0.2, 0) |- ($(Generator)-(0.5,0.14)$);

\begin{scope}[shift={(Generator)}]
    \fill[network, draw=black] (-0.5,-0.35) -- (0.5,-0.6) -- (0.5,0.6) -- (-0.5,0.35) -- cycle;
    \node[draw=none, anchor=center] at (0,0) {G};
\end{scope}

\pic [fill = white!80!black, pic text = {$\bx_\text{gen}$}] at (xgen) {vecinp};

\draw[->] ($(Generator)+(0.5,0)$) -- ($(xgen)-(0.1,0)$);  
\draw ($(xgen)+(0.1,0)$) -| ($(pfake)-(0.55,0)$) |- (pfake);  

\draw[->] ($(xgen)+(0.1,0)$) -- ($(xgen)+(0.45,0)$) |- ($(ygen)-(0.1,0)$) node[draw=black, fill=white, pos=0.75, anchor=center, align=center] {extract from};

\pic[fill = label, pic text = {$\by_\text{gen}$}] at (ygen) {vecinp};

\draw[->] ($(ygen)+(0.1,0)$) -- +(0.71, 0) node[anchor=west, draw=black, minimum width=1cm] (regloss) {$\mathcal{L}_\text{reg}$};
\draw[<-] (regloss) -- ($(regloss)+(0,0.6)$) node[anchor=south] {$\by_\text{cond}$};

\coordinate (x) at ($(yhat)+(0,1.6)$);
\coordinate (y) at ($(x)+(0,1.2)$);

\pic [fill = white!80!black, pic text = {$\bx_\text{real}$}] at (x) {vecinp};
\pic [fill = label, pic text = {$\by_\text{real}$}] at (y) {vecinp};
\draw[dashed] ($(x)-(0.36,0.5)$) rectangle ($(y)+(0.36,0.8)$);
\node[rotate=90] at ($(x)!0.5!(y)-(0.6,0)$) {dataset};

\draw ($(x)+(0.1,0)$) -| ($(preal)-(0.55,0)$) |- (preal);

\begin{scope}[shift={(Discriminator)}]
    \fill[network, draw=black] (-0.5,-0.6) -- (0.5,-0.35) -- (0.5,0.35) -- (-0.5,0.6) -- cycle;
    \node[draw=none, anchor=center] at (0,0) {D};
\end{scope}

\draw[->] ($(Discriminator)+(0.5,0)$) -- +(1.2, 0) node[anchor=north, pos=0.5] {$D(\bx)$}  node[anchor=west, draw=black, minimum width=1cm] {$\mathcal{L}_\text{adv}$};

\end{tikzpicture}}
	\caption{
		\textbf{cGAN with additional regression}:
		The generator~$G$ generates a point cloud~$\bx_\text{gen}$ from a random vector~$z$ and a regression label~$y$.
		The discriminator either receives a real or a generated point cloud and predicts the probability of the sample originating from the real distribution.
		Additionally, the dimensions~$\by_\text{gen}$ of the generated point cloud are extracted which are then used to compute the regression error~$\mathcal{L}_\text{reg}$.
	}
	\label{fig:reggan}
\end{figure}
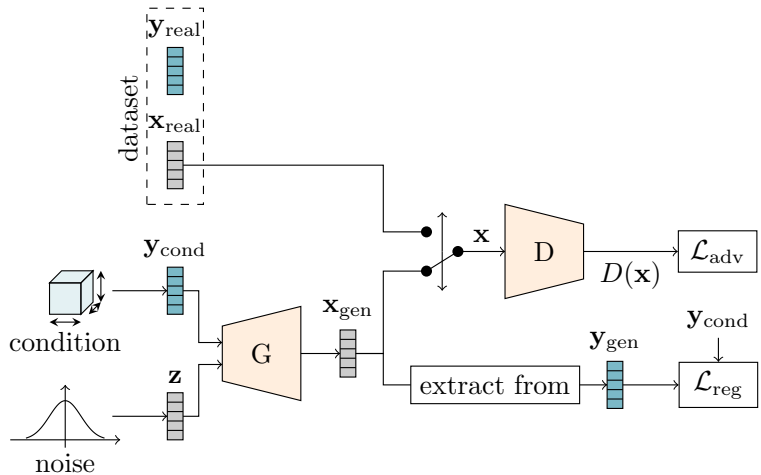

The application of influencing the dimensions of an object has the major advantage that it is possible to directly compute the dimensions from the generated object to check whether the generator worked properly.
This can also be used as a training signal.
\figref{fig:reggan}~shows an alternative approach for our proposed architecture, where a standard unconditioned discriminator is combined with an additional regression component.
The generation of the desired dimensions is explicitly enforced with the regression loss, therefore we refer to this variant as the Regression cGAN.
The generator loss is defined as
\begin{equation*}
    \mathcal{L}_G =
    - \nE_{\bz \sim \mathcal{Z}} \left[  D\left( G(\by_\text{cond}, \bz) \right) \right]
    + \lambda_\text{reg} \cdot \mathcal{L}_\text{reg} \left( \by_\text{cond}, \by_\text{gen} \right)
\end{equation*}
where $\by_\text{gen}$ are the actual dimensions calculated from the generated point cloud, and $\lambda_\text{reg}$ is the weighting factor for the regression loss.
The generator loss is solely responsible to enforce the adherence of the dimension conditioning since the discriminator is not conditioned on the label input.
The discriminator only judges from which distribution a sample originates from.
Its loss function is defined as
\begin{equation*}
    \mathcal{L}_D =
    \nE_{\bz \sim \mathcal{Z}} \left[ D\left(G(\by_\text{cond}, \bz) \right) \right]
    - \nE_{\bx \sim \mathcal{R}} \left[ D(\bx_\text{real}) \right]
    + \mathcal{L}_\text{gp}
    .
\end{equation*}

\begin{table}
    \centering
    \caption{
        \textbf{Quantitative Comparison}:
        We report results for the classes ``Chair'' and ``Airplane''.
        All evaluations are conducted on a hold-out validation split.
        For both \ac{fpd} and \ac{mse} smaller values indicate a better performance.
        \ac{mse} is given in \%.
    }
    \label{tab:reggan_results}
    \begin{tabularx}{0.5\linewidth}{X cc c cc}
        \toprule
        \multirow{2}{*}{Model} & \multicolumn{2}{c}{Chair} && \multicolumn{2}{c}{Airplane} \\
        \cline{2-3}\cline{5-6}
         & FPD & MSE && FPD & MSE \\
        \midrule
        Reg. cGAN & \textbf{1.4420} & 1.85 && 0.8802 & 20.19 \\
        Ours & 1.5290 & \textbf{0.28} && \textbf{0.8691} & \textbf{0.30} \\

        \bottomrule
    \end{tabularx}
\end{table}

The training behavior of the Regression cGAN is fundamentally different to the one of our proposed method.
We observe that \ac{fpd} is optimized first, hitting a minimum value at a quite early point during training where \ac{mse} is still quite high.
The results for this checkpoint are listed in~\tabref{tab:reggan_results}.
From this point onward \ac{mse} is further minimized at the expense of \ac{fpd} performance.
In contrast to our proposed method, Regression cGAN aims at minimizing \ac{mse} down to zero while accepting \ac{fpd} values that are magnitudes higher than for our proposed method.
This means no realistic object shapes are being generated.

Considering these results and the fact that Regression cGAN can only be used for specific applications where $\by_\text{gen}$ can easily be retrieved from the generated data, we found that our proposed method is superior to Regression cGAN.
Our method offers a much easier and more stable handling of training while resulting in a very good performing model that is applicable to many scenarios.

\section{ADDITIONAL RESULTS}
\label{suppl:results}

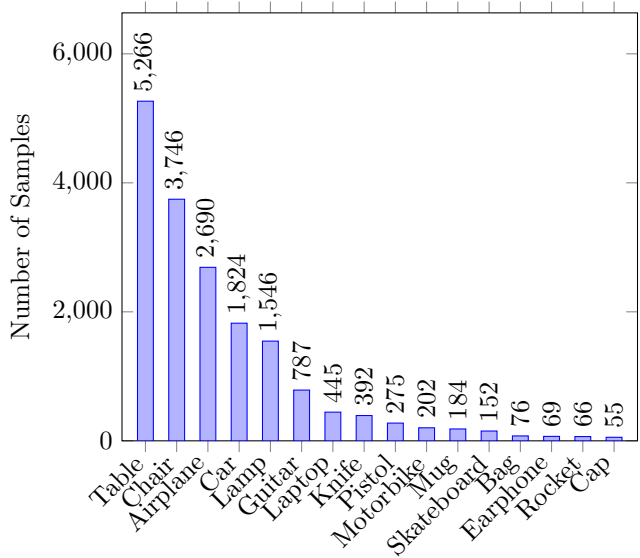
\begin{figure}
	\centering

	\pgfplotstableread[col sep=comma]{data/dataset.csv}\datatable

	\begin{tikzpicture}
		\begin{axis}[
			ybar,
			bar width=6pt,
			xtick=data,
			xticklabels from table={\datatable}{name},
			x tick label style={rotate=45,anchor=east},
			ymin=0,
			ylabel={Number of Samples},
			nodes near coords,
			every node near coord/.append style={black, rotate=90, anchor=west},
			enlarge y limits={upper,value=0.26},
			enlarge x limits=0.05,
			]

			\addplot table[x expr=\coordindex, y=count, col sep=comma] {data/dataset.csv};

		\end{axis}
	\end{tikzpicture}

	\caption{
		\textbf{Dataset Class Distribution}:
		Shown is the class distribution of the ShapeNetPart dataset~\citep{ShapeNetPart2016}.
		We use the five largest classes for our experiments: ``Table'', ``Chair'', ``Airplane'', ``Car'', and ``Lamp''.
	}
	\label{fig:dataset_classes}
\end{figure}

\begin{figure*}
	\centering

    \begin{subfigure}{0.19\linewidth}
        \resizebox{\linewidth}{!}{
            \begin{tikzpicture}
                \begin{axis}[
                    enlargelimits=false,
                    axis on top,
                    axis equal image,
                    xlabel=Object Width ($\Delta y$),
                    ylabel=Object Height ($\Delta z$),
                ]

                    \addplot graphics [xmin=0,xmax=1,ymin=0,ymax=1] {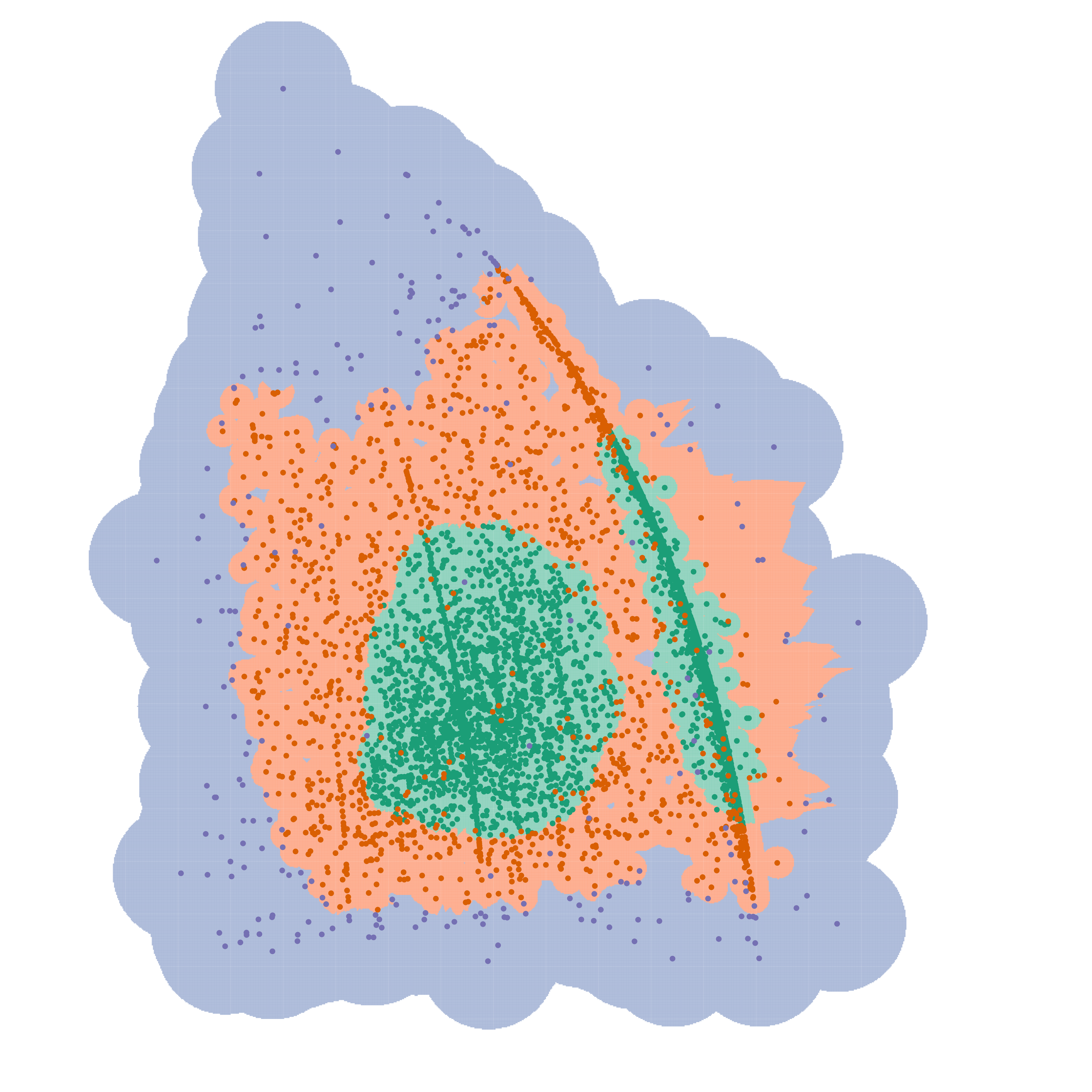};

                \end{axis}
            \end{tikzpicture}
        }
        \caption{Table}
    \end{subfigure}
    \begin{subfigure}{0.19\linewidth}
        \resizebox{\linewidth}{!}{
            \begin{tikzpicture}
                \begin{axis}[
                    enlargelimits=false,
                    axis on top,
                    axis equal image,
                    xlabel=Object Width ($\Delta y$),
                    ylabel=Object Height ($\Delta z$),
                ]

                    \addplot graphics [xmin=0,xmax=1,ymin=0,ymax=1] {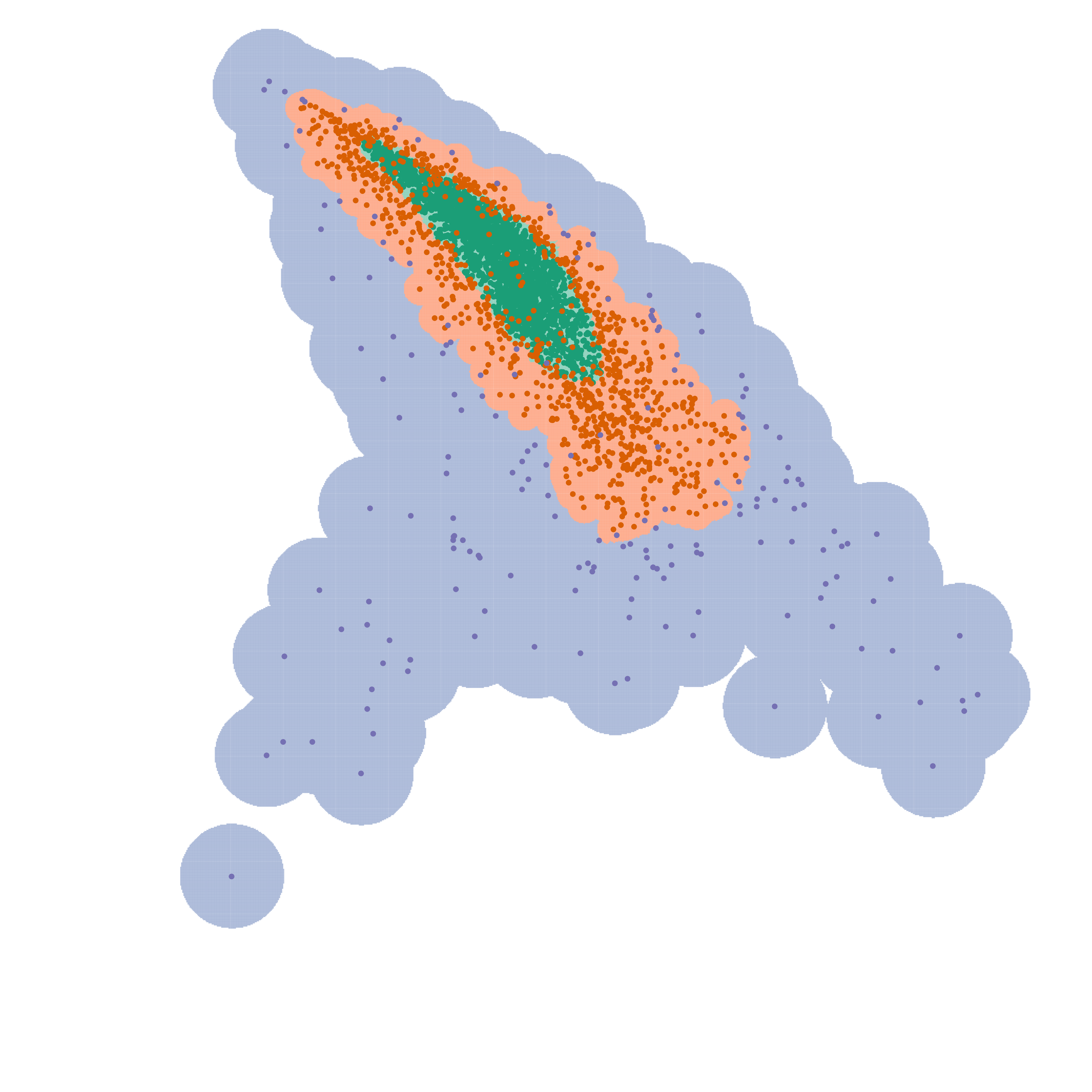};

                \end{axis}
            \end{tikzpicture}
        }
        \caption{Chair}
    \end{subfigure}
    \begin{subfigure}{0.19\linewidth}
        \resizebox{\linewidth}{!}{
            \begin{tikzpicture}
                \begin{axis}[
                    enlargelimits=false,
                    axis on top,
                    axis equal image,
                    xlabel=Object Width ($\Delta y$),
                    ylabel=Object Height ($\Delta z$),
                ]

                    \addplot graphics [xmin=0,xmax=1,ymin=0,ymax=1] {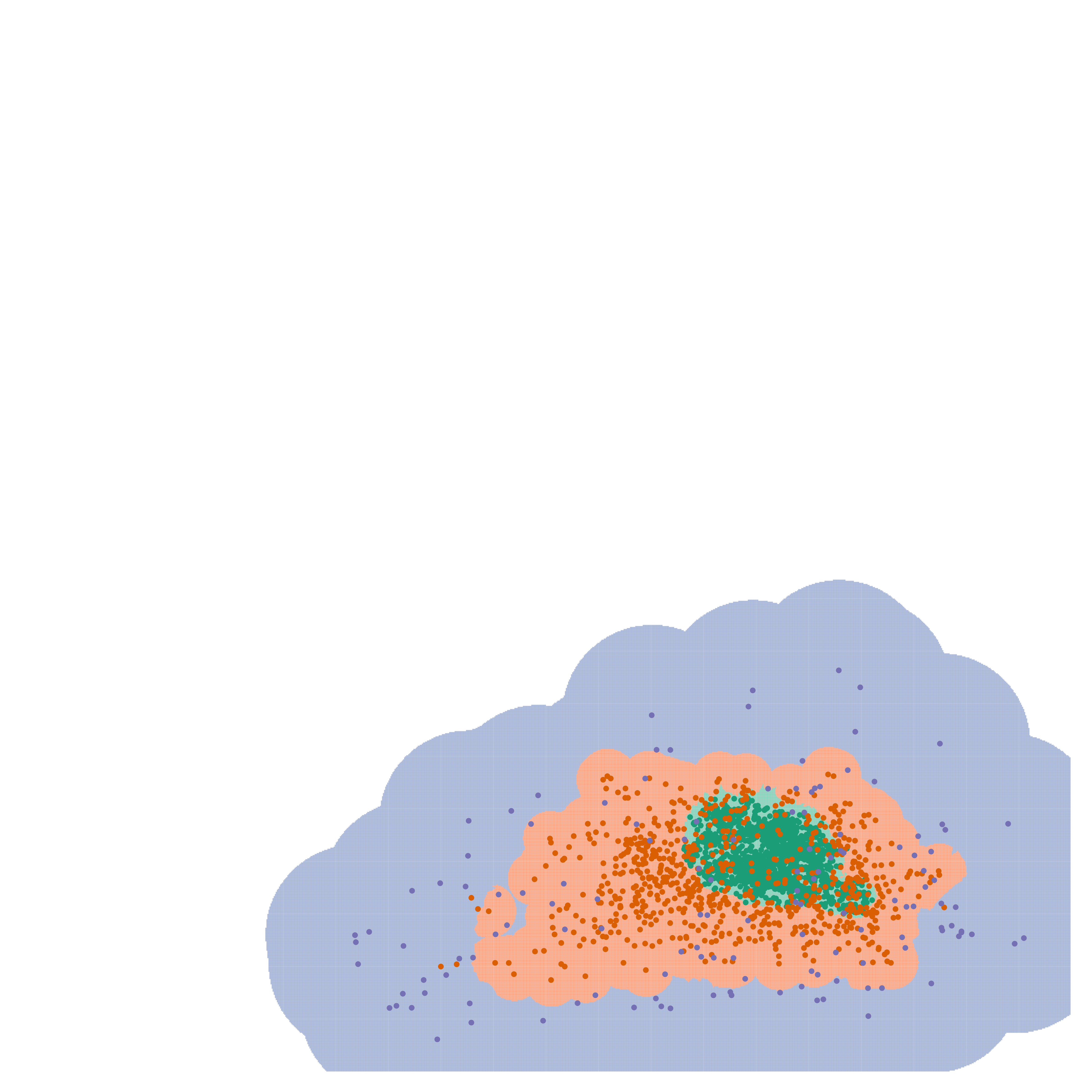};

                \end{axis}
            \end{tikzpicture}
        }
        \caption{Airplane}
    \end{subfigure}
    \begin{subfigure}{0.19\linewidth}
        \resizebox{\linewidth}{!}{
            \begin{tikzpicture}
                \begin{axis}[
                    enlargelimits=false,
                    axis on top,
                    axis equal image,
                    xlabel=Object Width ($\Delta y$),
                    ylabel=Object Height ($\Delta z$),
                ]

                    \addplot graphics [xmin=0,xmax=1,ymin=0,ymax=1] {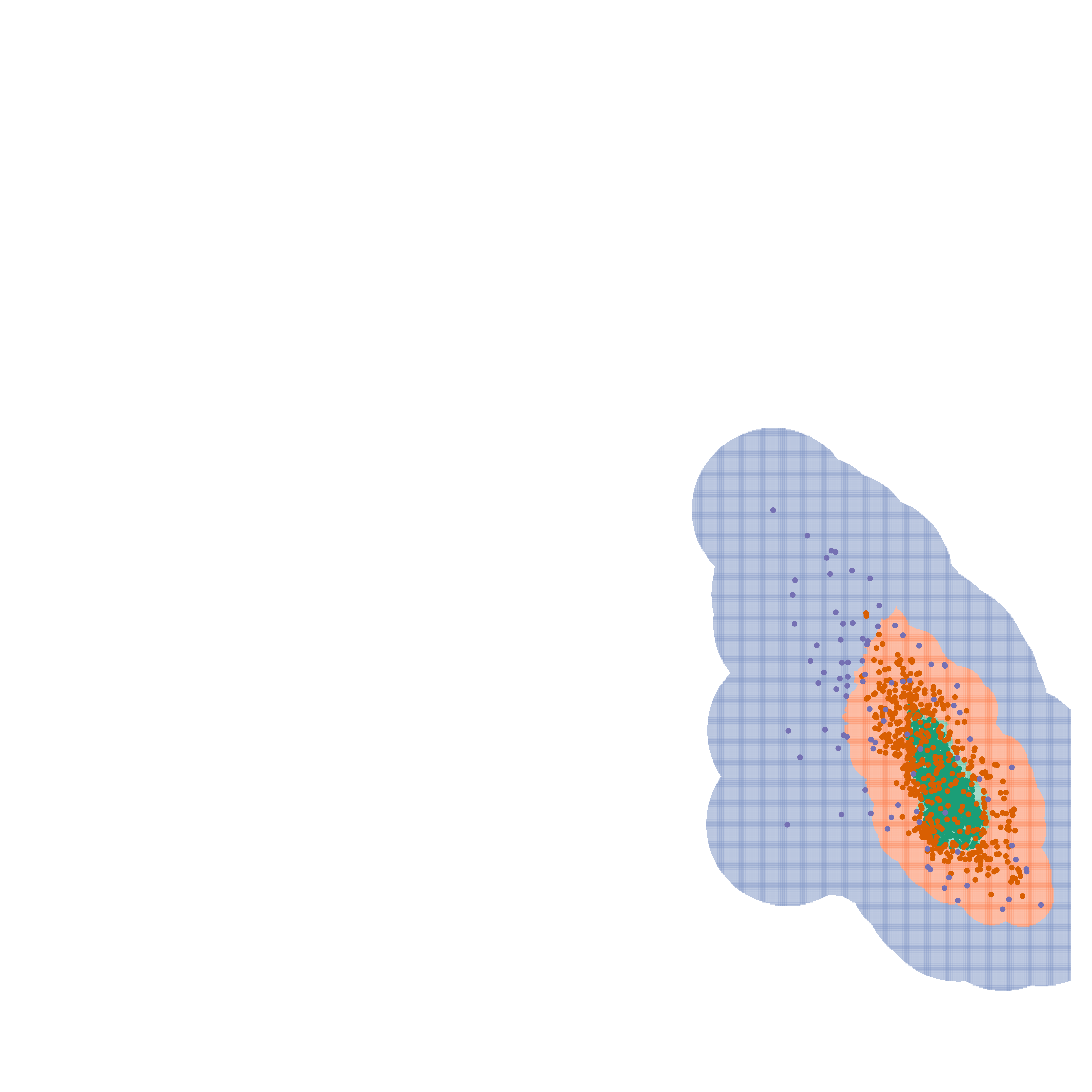};

                \end{axis}
            \end{tikzpicture}
        }
        \caption{Car}
    \end{subfigure}
    \begin{subfigure}{0.19\linewidth}
        \resizebox{\linewidth}{!}{
            \begin{tikzpicture}
                \begin{axis}[
                    enlargelimits=false,
                    axis on top,
                    axis equal image,
                    xlabel=Object Width ($\Delta y$),
                    ylabel=Object Height ($\Delta z$),
                ]

                    \addplot graphics [xmin=0,xmax=1,ymin=0,ymax=1] {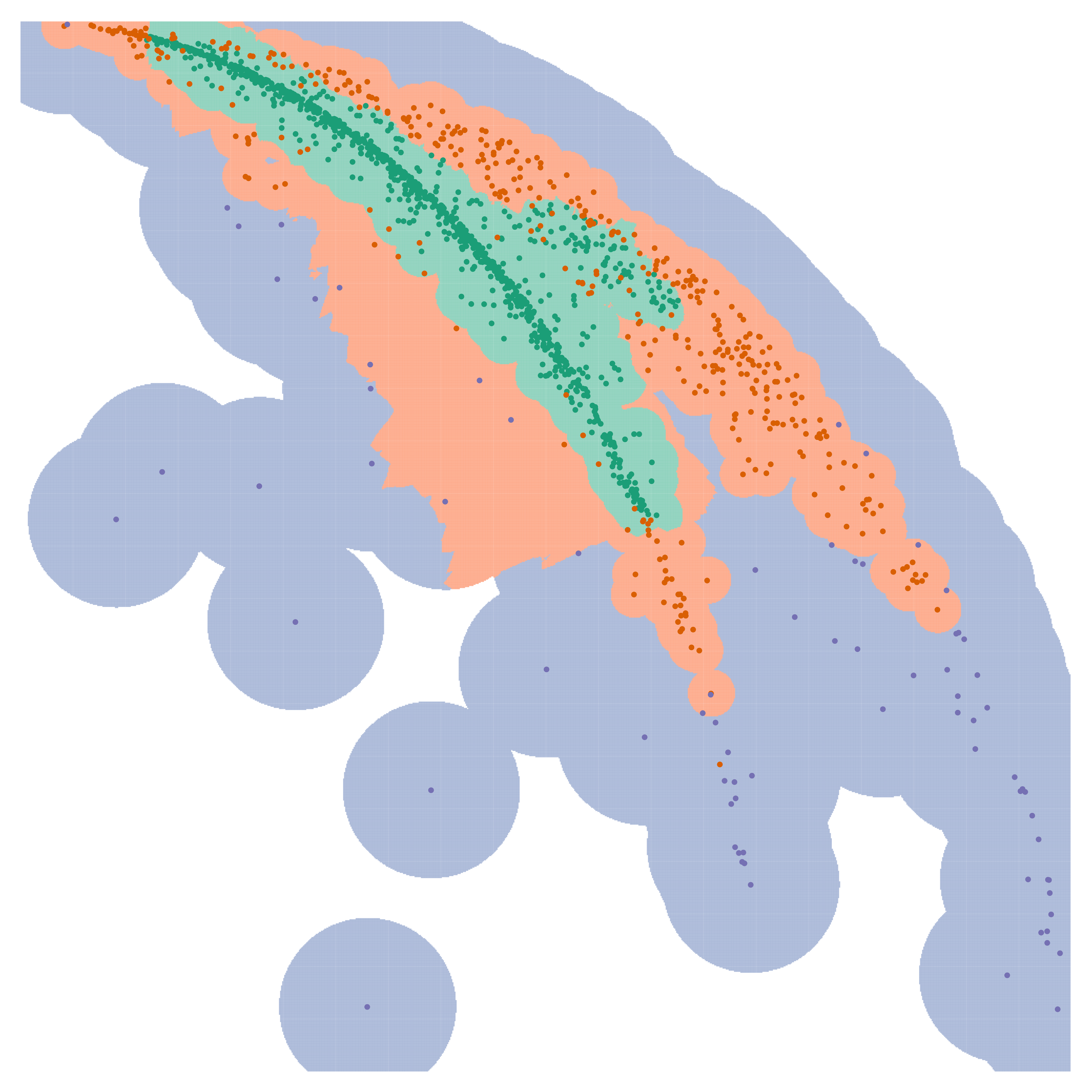};

                \end{axis}
            \end{tikzpicture}
        }
        \caption{Lamp}
    \end{subfigure}

	\caption{
		\textbf{Region-classified dimension distribution}:
		Shown is the sample distribution of five different object classes according to their extent in height~$z$ and width~$y$.
		The length~$x$ of the object is not considered in this visualization.
		Each mark corresponds to one shape in the dataset.
		The three colors represent the resulting regions from \ac{knn} classifier with $k$=20 based on a \ac{kde}.
		The regions correspond to $1\sigma\approx68\%$ ({\color{sigma1_d}green}), $2\sigma\approx27\%$ ({\color{sigma2_d}orange}), and $3\sigma=5\%$ ({\color{sigma3_d}blue}) of the entire data distribution.
		(Best viewed in color)
	}
	\label{fig:suppl_dimension_distribution}
\end{figure*}

\tabref{tab:quantitative_results}~displays the quantitative results of the five largest classes of the ShapeNetPart dataset and \tabref{tab:sigma_results} gives details on the region-based performance.
The class distribution of the dataset is shown in~\figref{fig:dataset_classes}.
For our experiments we did not consider classes with less than $1,000$ object shapes.
The quantitative results, especially when comparing our method to the chosen baselines, also depend on the distribution of the parameter we influence.
For comparison, \figref{fig:suppl_dimension_distribution}~shows the distribution of the five largest classes in terms of their extent in object height and width.
While some classes, like ``Table'' have a wide distribution, others are more densely packed, like ``Car''.
Especially for the ``Lamp'' class, the stretching baseline~(B2) achieves bad performance in terms of FPD, which can be attributed to its unique distribution.

\begin{table*}
    \centering
    \caption{
        \textbf{Quantitative Comparison}:
        We report results for the five largest classes of the ShapeNetPart dataset in terms of the metrics used by \cite{Shu2019ICCV} and~\cite{Achlioptas2018ICML}.
        Additionally, we report the regression error (MSE) for our introduced task.
        We freshly trained TreeGAN~\citep{Shu2019ICCV} to serve as our backbone network and report the results for reference.
        The baselines B1 and B2 are described in Sec.~5.3 of the main paper.
        Baseline~1 is the backbone network where ten versions per object are samples and the one with the best matching dimensions is chosen.
        Baseline~2 uses the backbone and then scales the object to the desired dimensions.
        All evaluations are conducted on a hold-out validation split.
        \cite{Shu2019ICCV}~use the entire dataset for training, therefore values might vary slightly.
    }
    \label{tab:quantitative_results}
    \begin{tabularx}{\linewidth}{X X c c cc c cc c}
        \toprule

        \multirow{2}{*}{Shape} &
        \multirow{2}{*}{Model} &
        \multirow{2}{*}{FPD ($\downarrow$)} &
        \multirow{2}{*}{MSE [\%] ($\downarrow$)} &
        \multicolumn{2}{c}{MMD ($\downarrow$)} &&
        \multicolumn{2}{c}{COV ($\uparrow$)} &
        \multirow{2}{*}{JSD ($\downarrow$)} \\
        \cline{5-6}\cline{8-9}
        & & & & CD & EMD && CD & EMD & \\

        \midrule

        \multirow{4}{*}{Table}
        & Backbone      & ~4.5245 &    -- & 0.0023 & 0.0863 && 0.4750 & 0.3750 & 0.1671 \\
        \cline{2-10}
        & Baseline~1    & ~3.3009 & 52.88 & 0.0026 & 0.0912 && 0.4875 & 0.3500 & 0.1423 \\
        & Baseline~2    & ~4.2851 & ~0.00 & 0.0028 & 0.0920 && 0.5375 & 0.3125 & 0.1661 \\
        & Ours          & ~3.0692 & ~0.16 & 0.0019 & 0.1073 && 0.4875 & 0.3000 & 0.1313 \\

        \midrule

        \multirow{4}{*}{Chair}
        & Backbone      & ~0.9525 &    -- & 0.0020 & 0.1027 && 0.4875 & 0.2500 & 0.1082 \\
        \cline{2-10}
        & Baseline~1    & ~1.3674 & 26.07 & 0.0023 & 0.1013 && 0.4750 & 0.2625 & 0.1123  \\
        & Baseline~2    & ~1.9259 & ~0.00 & 0.0021 & 0.1003 && 0.4875 & 0.2625 & 0.1068 \\
        & Ours          & ~1.5290 & ~0.28 & 0.0022 & 0.1059 && 0.4625 & 0.3125 & 0.1434 \\

        \midrule

        \multirow{4}{*}{Airplane}
        & Backbone      & ~1.2947 &    -- & 0.0002 & 0.0805 && 0.4375 & 0.1375 & 0.1887 \\
        \cline{2-10}
        & Baseline~1    & ~1.0209 & 15.08 & 0.0003 & 0.0812 && 0.4500 & 0.1125 & 0.1819 \\
        & Baseline~2    & ~1.6613 & ~0.00 & 0.0003 & 0.0783 && 0.5250 & 0.1375 & 0.1834 \\
        & Ours          & ~0.8691 & ~0.30 & 0.0003 & 0.0724 && 0.5000 & 0.1250 & 0.1291 \\

        \midrule

        \multirow{4}{*}{Car}
        & Backbone      & ~1.0816 &   -- & 0.0009 & 0.0656 && 0.4250 & 0.2375 & 0.0692 \\
        \cline{2-10}
        & Baseline~1    & ~2.6293 & ~5.15 & 0.0009 & 0.0651 && 0.4500 & 0.2000 & 0.0743 \\
        & Baseline~2    & ~2.2045 & ~0.00 & 0.0009 & 0.0634 && 0.4375 & 0.2750 & 0.0670 \\
        & Ours          & ~1.7129 & ~0.69 & 0.0010 & 0.0708 && 0.4125 & 0.1250 & 0.0714 \\

        \midrule

        \multirow{4}{*}{Lamp}
        & Backbone      & ~2.9954 &    -- & 0.0037 & 0.1630 && 0.4500 & 0.2750 & 0.2642 \\
        \cline{2-10}
        & Baseline~1    & ~3.4737 & 79.65 & 0.0029 & 0.1614 && 0.4500 & 0.2125 & 0.2569 \\
        & Baseline~2    & 36.5025 & ~0.00 & 0.0041 & 0.1606 && 0.4500 & 0.2250 & 0.2653 \\
        & Ours          & ~7.5012 & ~0.77 & 0.0038 & 0.1917 && 0.4375 & 0.1750 & 0.2576 \\

        \midrule
        \midrule

        \multirow{4}{*}{\textbf{Total}}
        & Backbone      & 2.1697 &    -- & 0.0018 & 0.0996 && 0.4550 & 0.2550 & 0.1595 \\
        \cline{2-10}
        & Baseline~1    & \textbf{2.3584} & 35.77 & \textbf{0.0018} & 0.1000 && 0.4625 & 0.2275 & 0.1535 \\
        & Baseline~2    & 9.3159 & \textbf{~0.00} & 0.0020 & \textbf{0.0989} && \textbf{0.4875} & \textbf{0.2425} & 0.1577 \\
        & Ours          & 2.9363 & ~0.44 & \textbf{0.0018} & 0.1096 && 0.4200 & 0.2075 & \textbf{0.1466} \\

        \bottomrule
    \end{tabularx}
\end{table*}
\begin{table*}
    \centering
    \caption{
        \textbf{Region-based performance}:
        The table shows performances for three sampling regions ({\color{sigma1_d}$\sigma_1$}, {\color{sigma2_d}$\sigma_2$}, {\color{sigma3_d}$\sigma_3$}) of the data distribution for our architecture.
        The model is trained with two different sampling strategies each: our proposed version where labels are sampled from the KDE of the distributions (areas of \figref{fig:dimension_distribution}), and the default version where labels are sampled from the list of labels contained in the dataset (marks of~\figref{fig:dimension_distribution}).
        For each $\sigma$-region, 1000 samples are generated.
        We report the MSE of the dimension regression and the FPD of the generated samples.
        Especially for less densely populated regions, \ie $\sigma_3$, our sampling strategy achieves better results.
    }
    \label{tab:sigma_results}
    \begin{tabularx}{\linewidth}{X p{5cm} ccc c ccc}
        \toprule

        \multirow{2}{*}{Shape} & \multirow{2}{*}{Label Sampling} & \multicolumn{3}{c}{FPD ($\downarrow$)} && \multicolumn{3}{c}{MSE [\%] ($\downarrow$)} \\
        \cline{3-5}\cline{7-9}
         & & $\sigma_1$ & $\sigma_2$ & $\sigma_3$ && $\sigma_1$ & $\sigma_2$ & $\sigma_3$ \\

        \midrule
        \multirow{2}{*}{Table}  & from distribution [ours] & {\bf4.7881} & {\bf11.1556} & {\bf83.1944} && {\bf0.14} & {\bf0.27} & {\bf17.49} \\
                                & from dataset samples & 5.5063 & 23.1117 & 153.7859 && 0.25 & 1.15 & 52.01 \\

        \midrule
        \multirow{2}{*}{Chair}  & from distribution [ours] & 3.1137 & {\bf2.4310} & {\bf18.3729} && 0.22 & {\bf0.26} & {\bf6.51} \\
                                & from dataset samples & {\bf2.5915} & 3.8257 & 138.0703 && {\bf0.13} & 0.41 & 25.30 \\

        \midrule
        \multirow{2}{*}{Airplane}   & from distribution [ours] & 2.8487 & {\bf1.6660} & {\bf22.4949} && 0.20 & {\bf0.36} & 7.62 \\
                                    & from dataset samples & {\bf2.5138} & 2.0912 & 42.4436 && {\bf0.17} & 0.39 & {\bf5.17} \\

        \midrule
        \multirow{2}{*}{Car}    & from distribution [ours] & 4.7105 & 11.8133 & {\bf32.1120} && {\bf0.61} & {\bf0.89} & {\bf3.74} \\
                                & from dataset samples & {\bf3.7981} & {\bf8.1593} & 60.2905 && 0.77 & 1.13 & 4.13 \\

        \midrule
        \multirow{2}{*}{Lamp}   & from distribution [ours] & 5.7873 & \textbf{20.2216} & \textbf{111.7557} && \textbf{0.44} & \textbf{1.84} & \textbf{11.50} \\
                                & from dataset samples & \textbf{4.4660} & 26.0117 & 290.8956 && 1.36 & 8.16 & 69.97 \\

        \bottomrule
    \end{tabularx}
\end{table*}

\figref{fig:random_samples} gives an overview on the quality and the diversity of the generated samples for the five different shapes.

\input{figures/suppl_random_samples}  

\end{document}